\documentclass{article}

\usepackage{microtype}
\usepackage{graphicx}
\usepackage{subfigure}
\usepackage{booktabs} 

\usepackage{hyperref}

\usepackage[accepted]{icml2024}

\usepackage{amsmath}
\usepackage{amssymb}
\usepackage{mathtools}
\usepackage{amsthm}

\usepackage[capitalize,noabbrev]{cleveref}

\usepackage{ulem}
\usepackage{url}
\usepackage{amsfonts}       
\usepackage{nicefrac}       
\usepackage{xcolor}         

\usepackage{wrapfig}

\theoremstyle{plain}

\theoremstyle{definition}

\theoremstyle{remark}

\usepackage[textsize=tiny]{todonotes}

\usepackage{mathrsfs}
\usepackage{amsthm}
\usepackage{amsmath,bm,bbm}
\usepackage{amsmath,amssymb,amsthm}
\usepackage{thmtools}
\usepackage{thm-restate}

\usepackage{multirow}
\usepackage{tabularx}
\usepackage{multicol}

\usepackage{amsfonts}
\usepackage{url}
\usepackage{mathtools}
\usepackage{float}
\usepackage{threeparttable}

\usepackage{algorithm}
\usepackage{algorithmic}
\renewcommand{\algorithmicrequire}{\textbf{Input:}}
\renewcommand{\algorithmicensure}{\textbf{Output:}}

\newcommand{\E}{\mathbb{E} }
\newcommand{\fig}{{Figure }}

\newcommand{\argmax}{{\arg\max }}

\newcommand{\alg}{{Algorithm }}

\newcommand{\ourmethod}{\text{Bilevel-LLM}}

\usepackage{balance}
\usepackage{tikz-cd}

\icmltitlerunning{Ask More, Know Better: Reinforce-Learned Prompt Questions for Decision Making with Large Language Models}
\begin{document}

\twocolumn[
\icmltitle{Ask More, Know Better: Reinforce-Learned Prompt Questions for Decision Making with Large Language Models}

\icmlsetsymbol{equal}{*}

\begin{icmlauthorlist}
\icmlauthor{Xue Yan}{yyy}
\icmlauthor{Yan Song}{yyy}
\icmlauthor{Xinyu Cui}{yyy}
\icmlauthor{Filippos Christianos}{sch}
\icmlauthor{Haifeng Zhang}{yyy}
\icmlauthor{David Mguni}{mar}
\icmlauthor{Jun Wang}{comp}
\end{icmlauthorlist}

\icmlaffiliation{yyy}{Institute of Automation, Chinese Academy of Science, Beijing, China}
\icmlaffiliation{comp}{University College London, UK}
\icmlaffiliation{sch}{Huawei Technologies, London, UK}
\icmlaffiliation{mar}{Queen Mary University, London, UK}

\icmlcorrespondingauthor{David Mguni}{davidmguni@hotmail.com}
\icmlcorrespondingauthor{Jun Wang}{jun.wang@cs.ucl.ac.uk}

\icmlkeywords{Machine Learning, ICML}

\vskip 0.3in
]




\printAffiliations{}

\begin{abstract}
Large language models (LLMs) demonstrate their promise in tackling complicated practical challenges by combining action-based policies with chain of thought (CoT) reasoning. Having high-quality prompts on hand, however, is vital to the framework's effectiveness. Currently, these prompts are handcrafted utilising extensive human labor, resulting in CoT policies that frequently fail to generalise. Human intervention is also required to develop grounding functions that ensure low-level controllers appropriately process CoT reasoning. 
{In this paper, we propose a comprehensive training framework for complex task-solving, incorporating human prior knowledge into the learning of action policies. 
}
To that purpose, we offer a new leader-follower bilevel framework that is capable of learning to ask relevant questions (prompts) and subsequently undertaking reasoning to guide the learning of actions. 
{The prompt policy is employed to make introspective revisions based on historical findings, leading the CoT process to consider the anticipated goals and generate outputs that lead to decisive, high-performing actions. }
The action policy subsequently learns to comprehend and integrate the CoT outputs to take actions. Our empirical data reveal that our framework outperforms leading methods in $5$ decision-making tasks such as Overcooked and FourRoom. 

\end{abstract}
\color{black}

\section{Introduction}

Large language models (LLMs) with Chain-of-thought (CoT) prompts \citep{wei2022chain, wang2022self} have achieved impressive performance improvements for solving complex natural language processing (NLP) tasks. {Moreover, techniques such as reward incentives and tree search \citep{yao2022react, hao2023reasoning}  have enhanced the quality of Chain-of-Thought reasoning for addressing intricate tasks, ultimately inducing the step-by-step problem-solving process.}
With the {impressive} reasoning capabilities of CoT, the reasoning outputs of LLMs can be used to provide useful `thought' inputs to policies that perform tasks in practical environments. 
This involvement of CoT reasoning has given rise to the promise of unlocking the power of LLMs to be able to assist in performing complex automated reasoning and acting in real-world environments.

While LLMs such as ChatGPT possess a wealth of human knowledge, in general,{ current methods \citep{yao2023tree,hao2023reasoning} heavily depend on meticulously crafted prompts designed by humans for each specific task}. Moreover, the performance of CoT reasoning can be {sensitive to the quality of the prompt input --- poor prompts provided even to powerful LLMs are unlikely to generate useful CoT outputs}. Additionally, despite the obvious potential of using CoT reasoning for guiding a low-level control policy, human-intelligible CoT reasoning can often be ambiguous for a downstream control policy, such as a rule-based planning method \citep{zhang2023proagent,shah2023navigation} and an action policy implemented by a reinforcement learning (RL) algorithm \citep{carta2023grounding}. 
As such, a natural consideration is for the need to generate CoT outputs that are interpretable to the action policy and, provably reduce the uncertainty of the action policy. Therefore, the ambition of embedding {appropriately} CoT reasoning within a generalist artificial intelligence (AI) framework has produced a series of critical challenges that have yet to be fully resolved. 

{In this paper, we propose a fully unified decision-making framework that incorporates CoT reasoning triggered by task-relevant prompts into performing complex tasks}.
In order to achieve this goal, both the prompt design and {the action policy} to be executed have to be sufficiently flexible and useful so as to adapt to the current task at hand. Tackling this challenge necessitates learning both to generate appropriate questions (a.k.a. prompts) given environment observations as well as learning how to {take the subsequent CoT outputs into account to }perform actions. To this end, we introduce a decision-making framework  which learns to ask pertinent \textit{questions} or perform introspection, performs CoT reasoning and then learns to take the best actions in the environment. The first component of the framework is enacted by a \textit{prompt policy} that learns a suitable prompt question given {the overall task challenges and its observations of the environment}. 
These prompts serve as inputs to a \textit{CoT process}, allowing the framework to perform task-related and meaningful reasoning given the prompt. The CoT \textit{thoughts} are then integrated into the \textit{action policy}, which learns to find solutions to tasks that may require both interaction experience and human knowledge embedded in CoT reasoning to solve.

Learning how to generate in-demand prompts for the CoT process produces formidable challenges. One such challenge is to ensure that the resulting CoT thoughts enhance the performance of an action policy. {Departing from a fixed set of pre-selected (human-crafted or automatically generated by a powerful LLM) prompts, and then learning to find a useful prompt from these to be fed into the CoT process presents an important challenge.} Specifically, ensuring that the resulting CoT thoughts improve the performance of an action-policy that can solve the task. 
{We resolve this challenge by designing a \textit{leader-follower Bilevel} structure, called $\ourmethod$ and illustrated in \fig \ref{fig:illustration}, that generates mutually adaptive policies.}  Each policy is endowed with its own objective --- the prompt policy observes the effect of its prompt and subsequent CoT reasoning on the action policy and learns to generate useful prompts. {In particular, the prompts are chosen so as to minimise the uncertainty of the action policy i.e.  minimise the entropy of the action-policy.} The action policy, on the other hand, learns to maximise the environmental reward while taking into account the outputs of the CoT process. {Ultimately, the generated thoughts serve to learn a more effective action policy, providing additional information beyond the observation of the environment. These natural language insights embody human knowledge, reducing the need for redundant exploration compared to traditional RL algorithms, which typically require extensive exploration of specific environments to train a competent agent.}

In numerous task environments, expert prompt data for the task is available, such as a well-defined set of subtasks \citep{shah2023navigation}. 
Making use of this in \textit{decision-making} tasks requires prompts that induce CoT reasoning for performing desirable actions at \textit{each state}. 
Nevertheless, often, the information in expert prompt sets is not refined to capture useful specifics at the state level producing a challenge of how to select the appropriate prompt at a given state. In environments where such prompt candidates are not available, the challenge becomes autonomously generating useful prompts using only the environment observations. In Sec. \ref{sec:experiments}, we demonstrate \ourmethod~is capable of tackling each of these challenges. First, we demonstrate that \ourmethod~successfully learns to select, from a global set of candidate prompts, the best prompt for each state. We then demonstrate that in problem settings where prompt candidates are not available, \ourmethod~successfully generates desirable and high-quality prompts from GPT-3.5 based on the task description. 

The contributions of this paper can be summarised as follows:
\newline$\bullet$ A new framework for auto-generation of prompts for decision-making tasks. An integral component is a prompt-policy or \textit{prompt-generator} which is trained by our framework to generate prompts that induce low uncertainty in the action-policy which receives thoughts generated by CoT reasoning triggered by the prompts from the prompt-generator. Therefore, the prompt-generator (and hence CoT process) behaves adaptively toward the needs of the action-policy.
\newline$\bullet$ A chain-of-thought generation framework in which the thought output of the CoT process are used to guide a policy that takes actions within an environment in order to solve practical tasks. This leverages the benefits of natural language models and CoT reasoning that encapsulate worldly experience and the capacity for deductive reasoning while efficiently tuning the thought pipeline process by tuning the prompt policy.
\newline$\bullet$ Prompt-tuning plus learning of CoT output-based policy that acts in environment ({bilevel} framework). 
\newline$\bullet$ A new bilevel learning algorithm that uses natural language to guide what actions and finds prompts for this desired textual guidance.

    

\section{Problem Formulation}

In this setting, an agent aims to solve some task by performing a sequence of actions in an environment. Formally, the problem is described by a partially observable Markov decision process (POMDP), which is defined by the following tuple $\langle \mathcal{S},\mathcal{A}, P, \mathcal{O}, T,\mathcal{R},\gamma\rangle$, where {$\mathcal{S}$ is the finite set of environment states}, $\mathcal{A}$ is the set of actions for the agent, $P:\mathcal{S}\times\mathcal{A} \rightarrow \Delta(\mathcal{S})$ is the state transition kernel for the environment, {$\mathcal{O}$ is the finite set of observations. The states and observations can be represented as symbolic vectors, which can also be translated into text descriptions conveying the information in the vectors.} The function $\mathcal{R}:\mathcal{S}\times\mathcal{A} \rightarrow \mathbb{R}$ is the reward function, which returns a scalar reward conditioned on a state-action pair whose realisation at time step $t$ we denote by $r_t\sim R$ and $\gamma\in [0,1]$ is the
discount factor.
{Lastly, the observation function is $T:\mathcal{S}\times \mathcal{A}  \to \mathcal{O}$ which is a mapping from the environment state, action to the observation set of the agent.}
Since the exact form of the observation and state spaces varies between environments, we provided a general description of the POMDP setting for introducing the general problem setting. \footnote{In the MDP setting, the observation space is equivalent to the state space.} We introduce an additional variable $x_t\in X$ contained in the \textit{situation space} $X$. {The variable $x_t$ represents the observable primary information of the state $s_t$, which can be encoded as text and, consequently, be utilized to assist the CoT process in generating task-relevant thoughts.}
%
%
\begin{figure*}[t]
	\centering
  		\includegraphics[width=0.9 \linewidth]{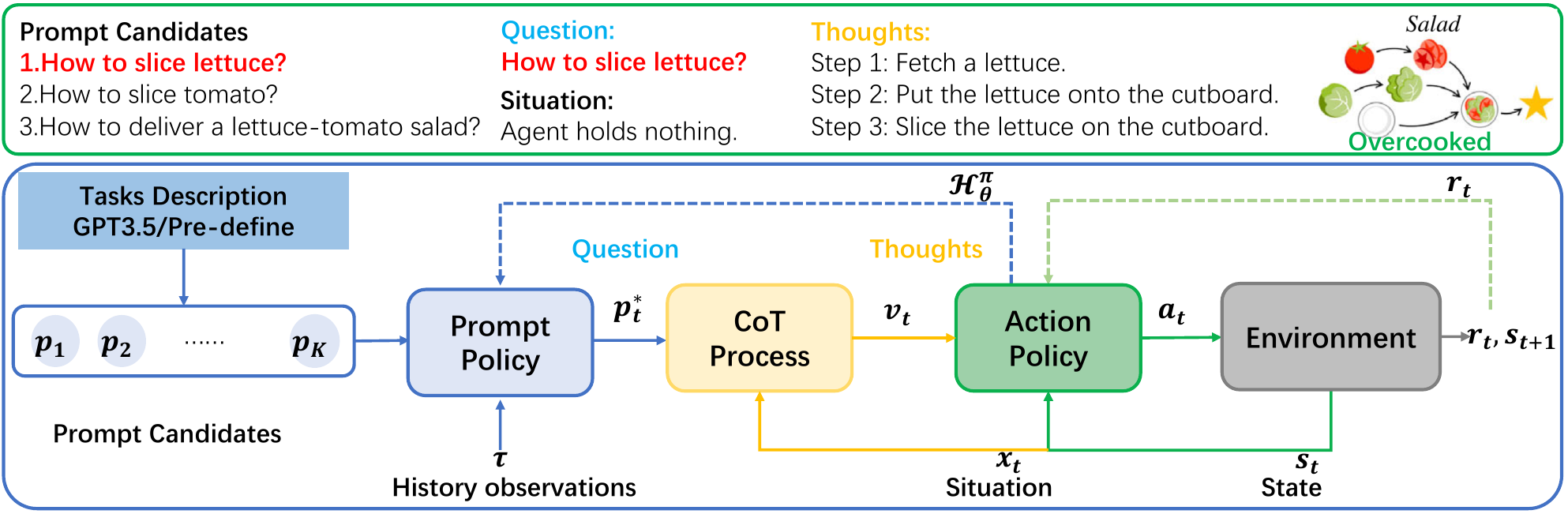}
       \caption{\textit{Top:} Example of the workflow from prompt candidates to CoT reasoning on Overcooked. 
       The prompt policy first selects a prompt question from the candidate set. Subsequently, the CoT process generates complex reasoning guided by the prompt and the current state situation to assist in the subsequent action performing. \textit{Bottom:} The illustration of our bilevel optimisation framework. }

    \label{fig:illustration}
\end{figure*}
In complex decision-making problems, standard methods such as RL struggle to solve these tasks in a sample efficient way. To solve these problems, an agent may required to deductive reasoning in order to resolve the challenge of finding an optimal policy. 

To tackle these challenges, we equip the agent with a {bilevel} LLM framework that enables the agent to first, generate its own pertinent prompts from its observations of the current, e.g. how to slice a lettuce on Overcooked task. Then, using these prompts, the CoT process performs complex reasoning about the best course of action, {e.g., the CoT outputs: fetching a lettuce, placing it onto the cutting board, and then slicing the lettuce on the board.} 
Lastly, an action policy takes an action according to the CoT reasoning and the observation in the environment. Examples of prompts questions and CoT reasoning are illustrated in \fig \ref{fig:illustration}.
The framework can therefore be split into three components: 
\newline
$\bullet$ First, a \textit{prompt policy} $\pi_\phi: (\mathcal{O})^{j<\infty} \rightarrow \boldsymbol{\mathcal{T}}$. {Denote that $\boldsymbol{\mathcal{T}}$ is the set of textual sentence based on the vocabulary set $\mathcal{V}$ (with finite words in it). Each sentence $t\in \boldsymbol{\mathcal{T}}\in \mathcal{V}^M$ with $M<\infty$ tokens in it.} This policy learns to generate prompts after observing (a window of) $j<\infty$ observations and outputs a thought in the textual thought space. 
\newline
$\bullet$ {Second, {a \textit{thought reasoning} policy $\pi^{\rm re}: X\times \boldsymbol{\mathcal{T}} \rightarrow \boldsymbol{\mathcal{T}}$}
--- an LLM that reasons about the task at the particular {state situation} and prompt question by performing CoT to generate a thought output.} 
The CoT reasoning is performed by a {fixed} LLM, since \ourmethod~is a plug \& play framework, any choice of LLM can be used to perform the CoT reasoning (in our experiments we use GPT-3.5 \footnote{The version of GPT-3.5 in this work is GPT-3.5-turbo}). 
\newline$\bullet$ {Lastly, an \textit{action-policy} $\pi_\theta: \mathcal{O}\times \boldsymbol{\mathcal{T}} \rightarrow \Delta(\mathcal{A})$. The action policy makes an observation of the environment and takes the CoT thought as an input then executes actions in the environment.} {Similarly, since \ourmethod~is a plug \& play framework and observations can be described using both textual and symbolic representations, the action policy can be served by an LLM with textual input or a traditional MLP with symbolic vector input.
In our experiments, we demonstrate the effectiveness of incorporating CoT reasoning into policy decision process under both settings.} 
 

%
%
%
%
%
%

%
%
%

{Concretely}, at times $t=0,1,\ldots$, a prompt $p_t$ is generated by the prompt policy i.e. $p_t\sim\pi_\phi(\cdot|o_{t},\ldots,o_{t-j\wedge 0})$. The prompt is then used by an LLM to trigger a CoT process whose output is a thought $\upsilon_{t}\in\boldsymbol{\mathcal{T}}$. Last, the action-agent samples an action from its policy $a_{t}\sim \pi_\theta(\cdot|o_{t}
,\upsilon_{t})$. 
Therefore, the sequence of events proceeds as follows:
\newline\textbf{1.} At time $t=0,1,\ldots$ the system is at an environment state $s_t\in \mathcal{S}$. 
\newline \textbf{2.} A prompt $p_t$ is produced by the prompt policy i.e. $p_t\sim\pi_\phi(\cdot|o_t,\ldots,o_{t-j\wedge 0})$. 
\newline \textbf{3.} An action $a_{t}\sim \pi_\theta(\cdot|o_t,\upsilon_{t})$ is taken given the output of the CoT process $\upsilon_{t}\sim\pi^{\rm re}(p_t,x_t)$, $x_t$ is the observed situation of the current state {(for example, the things that the agent current holds on the Overcooked)}. 
\newline \textbf{4.} The environment state transitions according to $s_{t+1}\sim P(\cdot|s_t,a_{t})$. 
\fig \ref{fig:illustrationprocess} in Appendix shows a step by step inference example of $\ourmethod$ on the Overcooked task.
\color{black}

To tackle the problem of learning how to generate prompts while learning the action-policy, we structure the problem as a leader-follower \textit{bilevel optimisation} \cite{colson2007overview}. This allows the prompt policy to learn how its decisions affect the action policy while the action policy and prompt policy learn concurrently. In this way, the prompt policy alters its output to guide the action policy in producing desirable actions while the action policy learns both how to interpret the CoT outputs and take desirable actions. Since LLMs already contain a vast amount of world knowledge, we here fix the LLM that performs the CoT reasoning, that is we assume that $\pi^{\rm re}$ is pretrained and fixed. We update the prompt policy and action policy. The aim of the prompt policy is to generate prompts minimise the uncertainty of the action policy. The optimisation objective can be expressed as a bilevel optimisation problem:
\begin{equation}    
\begin{split}
    (\pi^*_\theta,\pi^*_\phi) &\in \argmax_{(\pi_\theta,\pi_\phi)\in \Pi_\theta\times\Pi_\phi} \E_{\pi_\theta,\pi_\phi,\upsilon_t\sim\pi^{\rm re}}\\
&\left[-\sum_{t\geq 0}\gamma^t\mathcal{H}^{\pi_\theta}(y_{t})|y_{t}=(o_t,\upsilon_{t})\right] \nonumber\\
    \text{s.t.}\; \pi^*_\theta &\in \argmax_{\pi_\theta\in \Pi_\theta}\:\E_{\pi_\theta,p_t\sim\pi_\phi,\pi^{\rm re}}\left[\sum_{t\geq 0}\gamma_I^{t}r_{t} \right],\; \\
\end{split} 
\end{equation}
where $\mathcal{H}^{\pi_\theta}(y_{t})=\sum_{a_t\in\mathcal{A}}\pi_\theta(a_t|y_{t})\operatorname{log}\pi_\theta(a_t|y_{t})$, $y_{t}=(o_t,\upsilon_{t})$
which is the entropy of the policy $\pi_\theta$ and 
$\gamma_I,\gamma\in [0,1)$ are the discount factors for the action and prompt generation policies respectively and $r_t\sim \mathcal{R}$ is the environment reward. \color{black}
{Here, we explain the bilevel optimisation: \\ \textbf{In the inner loop}, the action policy $\pi_\theta$ learns to understand and incorporating the CoT reasoning containing high-level task-solving knowledge to perform optimally, i.e, maximising environment reward. \\
\textbf{In the outer loop}, the prompt policy $\pi_\phi$ aims to minimise the entropy of the action policy. It learns to generate appropriate prompts that subsequently leads CoT reasoning, enabling the action policy to take high-performing actions more certainly.
}
Note that the bilevel aspect incorporates the nested nature of the optimisation \cite{colson2007overview,dempe2002foundations} --- in order to find the optimal prompt, the prompt policy must take into account the anticipated behaviour of both the LLM $\pi^{\rm re}$ and the action policy $\pi_\theta$ and thereafter make its choice accordingly. 

\section{Methodology}
In this section, we describe the training procedure of the proposed Bilevel framework. The prompt policy is optimised via the policy gradient with the behavior of action policy as a reward. The action policy is optimised by PPO \cite{schulman2017proximal} updater, which benefits from avoiding human-crafted engineering when grounding the CoT reasoning to executable actions. In the bilevel framework, the prompt policy and action policy are concurrently optimised until convergence. The overall bilevel framework is illustrated in \fig \ref{fig:illustration}.

\begin{algorithm}[th!]
\caption{ $\ourmethod$}
\label{alg:bilevel}
\begin{algorithmic}[1]
    \REQUIRE Initialise parameters of policies $\pi_\theta$, $\pi_\phi$.  Set the data buffer $D=\emptyset$. 
    \ENSURE $\pi^*_\theta$, and $\pi^*_\phi$. 
    \WHILE{not done}
        \STATE \#Rollout trajectories with $\pi_\theta, \pi^{\rm re}, \pi_\phi$.
        \FOR{$i=1,2,..,\text{step}$}
            \STATE Generate prompt given historical observations: 
            $p_t \sim \pi_\phi(\cdot|o_{t},\ldots,o_{t-j\wedge 0}).$
            \STATE Perform CoT reasoning given prompt and state situation: $\upsilon_{t}\sim \pi^{\rm re}(\cdot|p_t,x_t)$.
            \STATE Sample action according to the CoT reasoning and observation:
             $a_{t} \sim \pi_\theta(\cdot|o_t,\upsilon_{t}).$
            \STATE 
            Apply action $a_{t}$ to the environment, sample the reward $r_{t}$ and next step observation $o_{t+1}$. 
            \STATE Calculate the entropy of the action policy: $h_{t}=\mathcal{H}\left(\pi_\theta(\cdot|s_t,\upsilon_{t})\right).$ 
            \STATE Add to data buffer: $D=D\cup{(o_t,p_t,\upsilon_{t},a_{t},r_{t},h_{t},o_{t+1})}$
        \ENDFOR
        \STATE Update the action policy $\pi_\theta$ via PPO to optimise Eq. (\ref{eq:ppo_objective}).
        \STATE Update the prompt policy $\pi_\phi$ by policy gradient following Eq. (\ref{eq:pg_gradient}).

        \color{black}
    \ENDWHILE
\end{algorithmic}
\end{algorithm}


%
\paragraph{prompt policy training via policy gradient.}
CoT reasoning with LLMs has proven to be effective in aiding decision-making when well-designed prompts are used \citep{zhang2023proagent, park2023generative}. However, the quality of CoT reasoning heavily depends on the quality of prompts, which are typically manually designed by humans \citep{zhang2023proagent, shah2023navigation}. In traditional Natural Language Processing (NLP) tasks such as sentiment classification \citep{pang2002thumbs} and news classification \citep{rana2014news}, prompts are usually provided through sets of input-output pairs, i.e., in-context learning examples. Unlike these NLP tasks with clearly defined input-output examples, the desired format of prompts varies across different decision-making tasks, often requiring substantial manual engineering.

{Due to the difficulty of training a model that is able to generate reasonable prompts from scratch automatically, we alternatively use pre-defined prompt candidates which can be obtained by human deliberately writing or being generated by GPT-3.5. In most of our experimental environments, we employ GPT-3.5 to generate a set of prompt candidates, where task descriptions are inputted into GPT-3.5 to produce task-relevant and critical prompt questions. 
{We also conduct an experiment to validate that, within our framework, the agent using prompt candidates generated by GPT-3.5 exhibits performance similar to that of the agent using prompts designed by humans. }}
Examples of prompt candidates generated from GPT-3.5 are shown in Appendix \ref{autopromptappendix}. 

{With a prompt candidate set $\mathcal{P}=\{p_1, p_2, \cdots p_K\}$, we train a prompt policy $\pi_\phi(\cdot|o_t,\ldots,o_{t-j\wedge 0})$ over the prompt candidates according to historical observations $o_t,\ldots,o_{t-j\wedge 0}$. 
Each of these natural language prompt candidates can be represented as a high-dimensional vector using a pre-trained and frozen Bert\cite{devlin2018bert} model. Denote the embedding of prompt candidate $p_i$ as $e_i$ and the embedding of the historical observations $(o_t,\ldots,o_{t-j\wedge 0})$ as $e_o=\mathcal{E}(o_t,\ldots,o_{t-j\wedge 0})$ with the encoder $\mathcal{E}$. The prompt embedding and the observation embedding are projected into the same vector space. Denote the mapped embedding as $\hat{e}_i=\mathcal{M}_p(e_i), \forall i=1\cdots K,\ \hat{e}_o=\mathcal{M}_o(e_o)$, where $\mathcal{M}_p$ and $\mathcal{M}_o$ are projectors. 
During the decision-making process, the prompt policy estimates the probability of selecting a prompt candidate $p_i$ based on the dot product and softmax operators, mathematically written as: ${\exp(\hat{e}_i^\top\hat{e}_o)}/{\sum_j\exp(\hat{e}_j^\top\hat{e}_o})$. The prompt policy is updated via the policy gradient with the minus entropy of the action policy as a reward incentive and parameters of the observation encoder $\mathcal{E}$ and projectors $\mathcal{M}_p$, $\mathcal{M}_o$ are trainable. }\color{black}
The detailed procedure is described as below:
\newline
    \newline$\bullet$ For a decision-making task, we employ GPT-3.5, along with the provided task description, to generate appropriate prompt candidates. As a second case, we used human-crafted assists to generate valuable prompt candidates.
    \newline$\bullet$ With these $K$ prompts, the prompt policy is optimised to maximise the minus action policy entropy. 
    The objective function is given by:
\begin{equation}    
\begin{split}
 \nonumber      &J_\phi(y|\pi_\theta,\pi_\phi,\pi^{\rm re})=\E_{\pi_\theta,\pi_\phi,\upsilon_{t}\sim\pi^{\rm re}}\left[-\sum_{t\geq 0}\gamma^t\mathcal{H}^{\pi_\theta}(y_{t})\right],
        \label{eq:pg_objective}
\end{split} 
\end{equation}
where $y_{t}=(o_t,\upsilon_{t})$ and $y_0=y$.
    \newline $\bullet$ We use the policy gradient \citep{lu2022dynamic} to optimise the prompt policy which obeys the following expression:
\begin{equation}
\begin{split}
    &\nabla_\phi {J}_\phi(y|\pi_\theta,\pi_\phi,\pi^{\rm re})
    \approx\\
    &\frac{1}{N} \sum_{t\geq0} \nabla_\phi\log\pi_\phi(p_t|o_t,\ldots,o_{t-j\wedge 0}) \hat{R}^o_t(\tau).
\label{eq:pg_gradient}
\end{split}
\end{equation}
$\hat{R}^o_t(\tau)= -\sum_{i\geq t}\left[\gamma^{i-t}\mathcal{H}^{\pi_\theta}(y_{i})|y_{i}=(o_i,\upsilon_{i}), \upsilon_{i}\sim\pi^{\rm re}\right]$ is the return-to-go from step $t$ to the end of a trajectory. The gradient is approximated by $N$ trajectories sampled from polices $\pi_\theta$, $\pi_\phi$ and $\pi^{\rm re}$. The detailed derivation of the gradient can be found in Appendix \ref{'appendix:gradient'}.
%

\noindent
\textbf{CoT reasoning with Prompts.} With the selected prompt $p_t$ sampled from the prompt candidate set, the CoT reasoning is obtained by $\upsilon_{t}\sim \pi^{\rm re}(\cdot|p_t,x_t)$, where the CoT process $\pi^{\rm re}$ is implemented by a powerful LLM such as GPT-3.5. The motivation of integrating the CoT reasoning into our bilevel framework to use the human prior knowledge to provide a high-level guideline for solving complicated decision-making tasks. {For example, as shown in \fig \ref{fig:illustration}, in Overcooked game, the CoT process can generate a sequence of crucial steps that need to be performed, prompted by a question related to a subtask ``how to slice lettuce''. }
To reduce the time and cost associated with frequent queries to GPT-3.5, we abstract situations to represent states and stored CoT outputs for the same situations. For example, in the case of two distinct states, even though the agent may be in different map positions and neither state involves holding lettuce, they are considered part of the same situation because the steps to slice lettuce remain the same: picking up a lettuce, placing it on the cutting board, and then proceeding to slice it. 

\noindent
\textbf{Action policy training via PPO with LLM.} Existing works~\citep{jang2021gpt,carta2023grounding} utilise LLMs as the action policy and fine-tune these LLMs to adapt to decision-making tasks, taking advantage of the comprehensive capabilities of LLMs. {In our work, we also utilise an LLM as the action policy. Within our framework, in addition to considering the textual observations provided by the environment, we also incorporate additional CoT reasoning from GPT-3.5 when performing actions. 
To regulate the action LLM to output executable actions, we fine-tune the action LLM, denoted as $\pi_\theta$, using Proximal Policy optimisation (PPO) \citep{schulman2017proximal}.} The objective of the action policy is to maximise the environment return:
\begin{equation}
\label{eq:ppo_objective}
    \arg\max_\theta \mathbb{E}_{a_t\sim\pi_\theta,p_t\sim\pi_\phi, \upsilon_t\sim\pi^{\rm re}}\left[\sum_{t\geq0} \gamma_I^t r_{t}\right]
\end{equation}
{We use the pre-trained LLM, Flan-T5 small \citep{rae2021scaling} with parameters less than one billion as the action policy.} {Our main method uses the LLM Flan-T5 small as action policy to better comprehend textual CoT thought. Additionally, our Plug \& Play framework also supports the use of an MLP network as the action policy with symbolic observations and CoT reasoning embedding as input. For more details, please refer to Section \ref{sec:experiments}.}


\noindent
\textbf{Bilevel Optimisation.}
In our leader-follower Bilevel LLM framework, the prompt policy and the action policies are trained alternately, with the other policy being kept frozen. On one hand, the prompt policy selects a prompt for the CoT process, the outputs of which are expected to assist the action policy in solving complex tasks. Thus, the goal of the prompt policy is to reduce the uncertainty of the action policy when it encounters challenging {scenarios}. In practical terms, the objective is to minimise the entropy of the action policy. On the other hand, the action policy is trained to effectively solve specific decision-making tasks while benefiting from CoT reasoning and the experience gathered during exploration. The overall training process of the Bilevel framework is detailed in \alg \ref{alg:bilevel}.

\section{Experiments}\label{sec:experiments}
In this section, We validate that our Bilevel framework, combining CoT reasoning and action policy learning, is beneficial for decision-making. Additionally, our bilevel framework supports learning prompt candidates from GPT-3.5 while also automatically interpreting CoT reasoning, thereby reducing the need for a large amount of human labor compared to LLM agents \cite{park2023generative,zhang2023proagent} that heavily rely on human annotation. 
Further details on experimental settings and more ablation study results can be found in the Appendix. 
\begin{figure*}[t!]

	\centering \includegraphics[width=0.75 \linewidth]{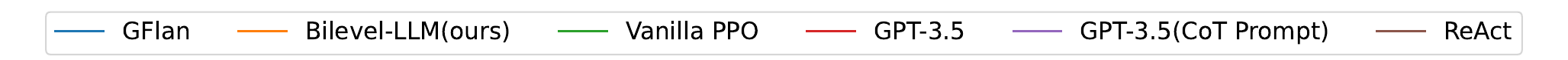}\\
{\includegraphics[width=0.20 \linewidth ]{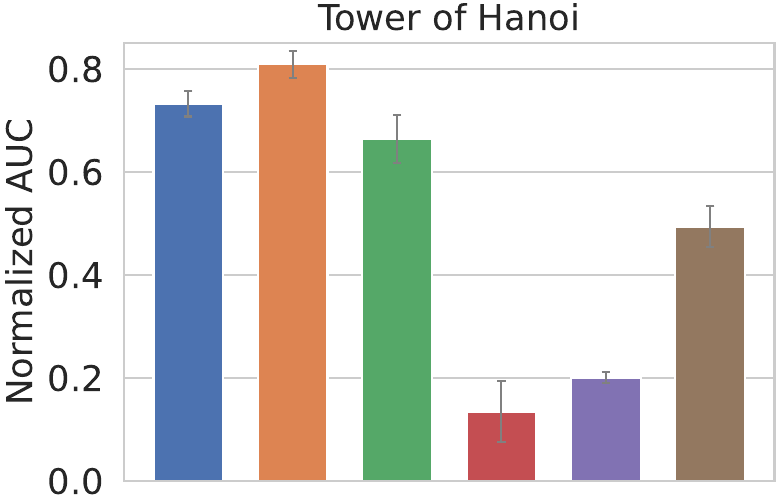}}
{\includegraphics[width=0.20 \linewidth ]{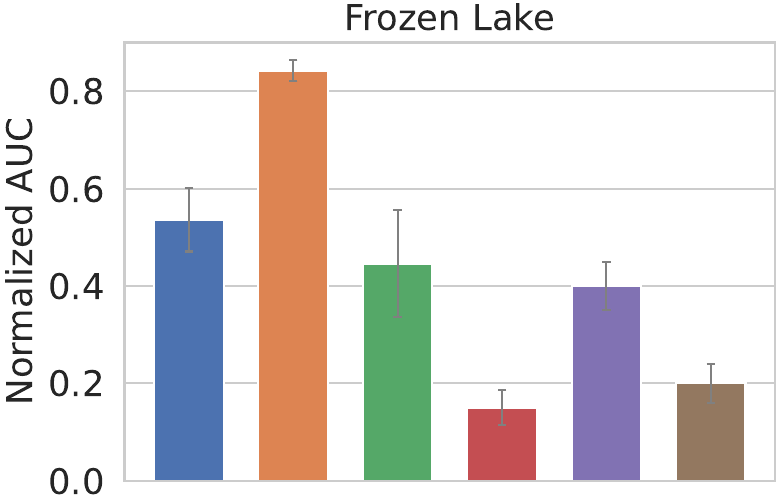}}
 {
    	\includegraphics[width=0.20 \linewidth]{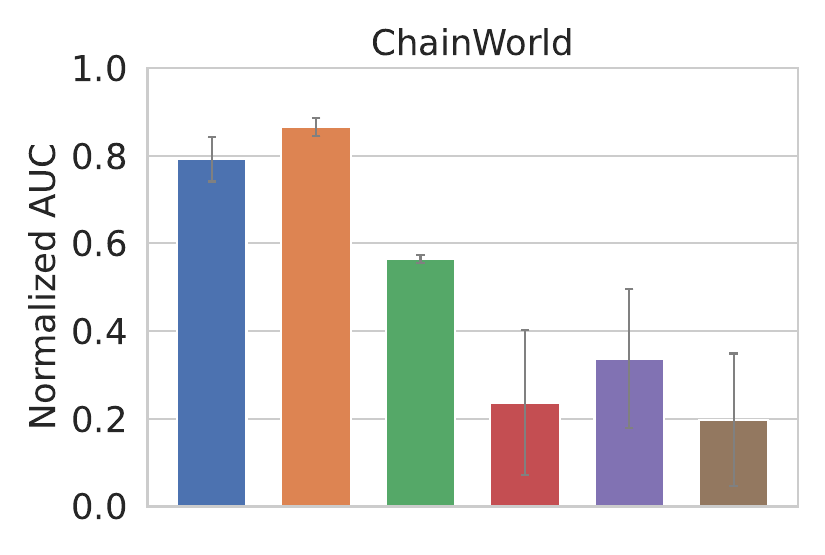}
    		}
    {    
    \includegraphics[width=0.20 \linewidth]{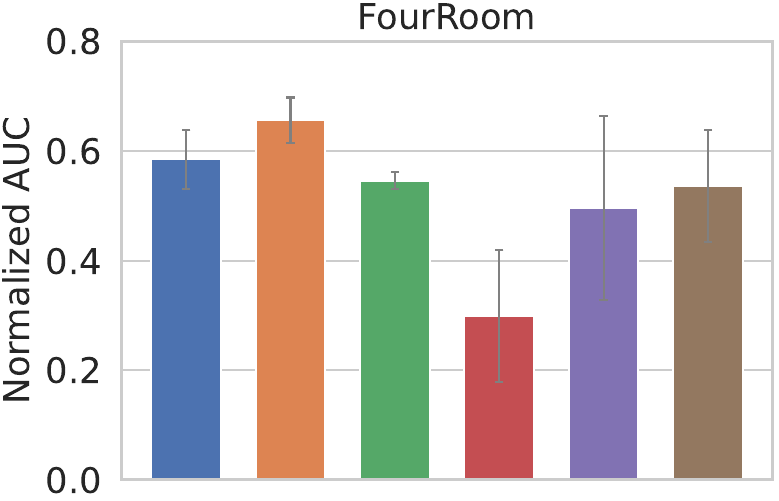}}
        {
    	\includegraphics[width=0.20 \linewidth]{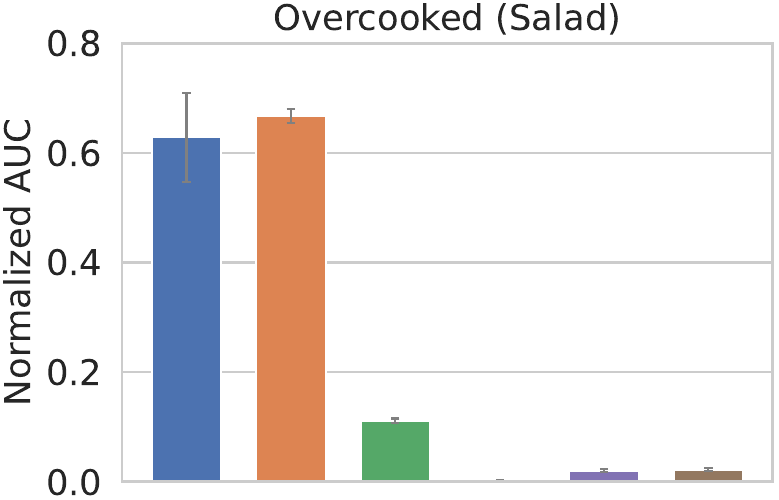}
    		}
        {
    \includegraphics[width=0.20 \linewidth]{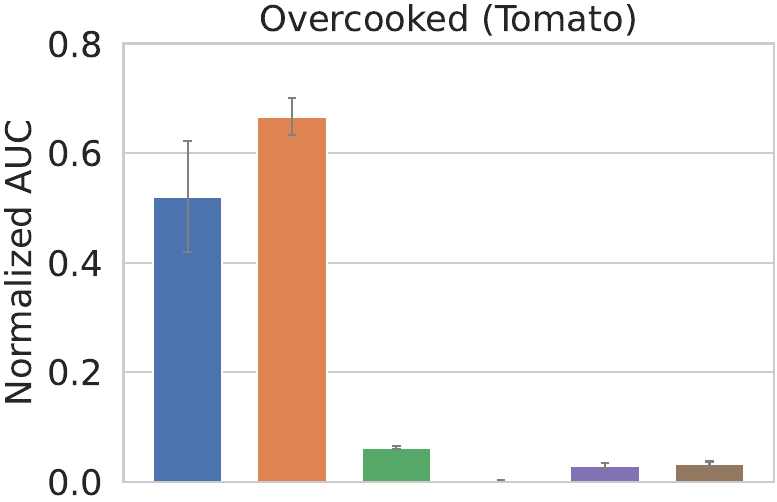}}
 {
    	\includegraphics[width=0.20\linewidth]{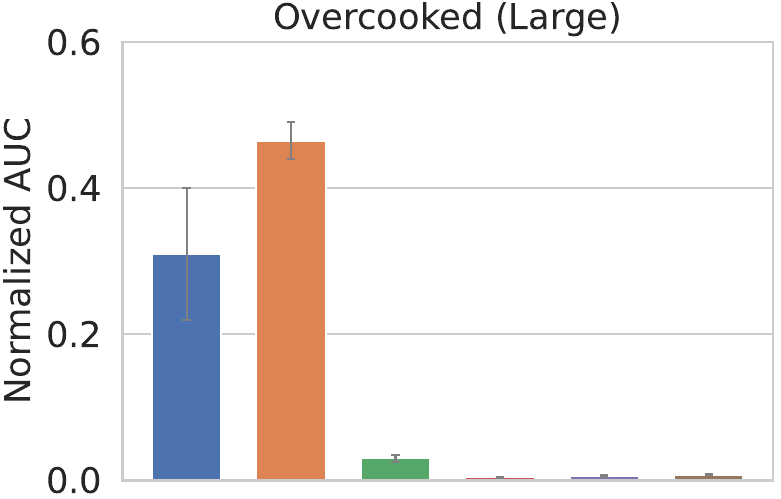}}
      \vspace{-0.7em}
    \caption{{Results of comparison with baselines. We plot the mean and standard error of nomalized reward averaged over $5$ seeds for trainable baselines, and over $20$ episodes for GPT-3.5 baselines. The cumulative rewards are normalized within the range $[0, 1]$, and the Area Under the Curve (AUC) is calculated by averaging over the entire training process. 
    }}
\label{fig:baselines}
\vspace{-1.0em}
\end{figure*}

\subsection{Environments}
We perform our empirical examinations on the following five environments: \textbf{Tower of Hanoi} \citep{hinz2013tower}, a classical logic reasoning game. \textbf{Frozen Lake} and textbf{ChainWorld} are POMDP environments, where only the agent's position is accessible. \textbf{FourRoom} is a POMDP task, where agent should navigate through hallways to reach the goal. On \textbf{Overcooked}, we consider three different layouts: \textit{Overcooked(Tomato)}: deliver a chopped tomato with the map size of $5\times4$; \textit{Overcooked(Salad)}: deliver a tomato-lettuce salad with a map size of $5\times4$; \textit{Overcooked(Large)}: deliver tomato-lettuce salad with a map size of $7 \times 7$. Note that the state space of \textit{Overcooked(Large)} reachs $9.8\times 10^{21}$, making it challenging to explore. More detailed environment description can be found in Appendix \ref{envs}.

\subsection{Baselines}
We compare $\ourmethod$ with two trainable baselines and three baselines that directly prompt GPT-3.5 to perform actions, namely:
\newline\textbf{GFlan} \citep{carta2023grounding}. GFlan adopts the LLM Flan-T5 large as the foundation of action policy and optimises it via PPO algorithm. GFlan solely relies on textual observations as input and employs this information to estimate the conditional probabilities of the action tokens.
\newline\textbf{Vanilla PPO} \citep{schulman2017proximal}. Unlike GFlan which leverages LLMs, Vanilla PPO employs MLP network as the backbone of the action policy and trains it from scratch. It use the symbolic embedding of states as the input of the action policy. 
\newline\textbf{GPT-3.5.} 
We test the zero-shot decision-making capability of GPT-3.5 with task descriptions, textual context, and executable action candidates as input prompt and let GPT-3.5 infer the action at the current state. 
\newline\textbf{GPT-3.5 with CoT prompt.} CoT prompts have the potential to substantially enhance the performance of GPT-3.5 on complex reasoning tasks. Besides the inputs used in the GPT-3.5 setting, we further incorporate examples of human interactions with the environment or human-established task decompositions as a part of the input prompt and instruct GPT-3.5 to think step by step. 
\newline\textbf{ReAct} \cite{yao2022react} This method leverages LLMs (GPT-3.5 turbo in our experiments) first to think the solution and then generate actions based on think. The ability to think and action reasoning is emergent via human-designed in-context examples. The method also keeps track of the historical thinking, action, and environment observation, and uses them as part of the prompt information for the next action. 
\newline\textbf{$\ourmethod$.} 
Motivated by the rich prior knowledge and textual comprehensive ability of pretrained LLM, our main method leverages the Flan-T5 small (with fewer than $1$ billions parameters) as the action policy, like GFlan. However, $\ourmethod$ combines CoT reasoning, revealing human-like high-level solution to tasks, to further improve the task-solving ability.


\begin{figure*}[t!]

	\centering

	\subfigure[Ablation on prompt policy]{
 \label{fig:ablationprompt}
    	\includegraphics[width=5.5cm ,height=2.0cm]{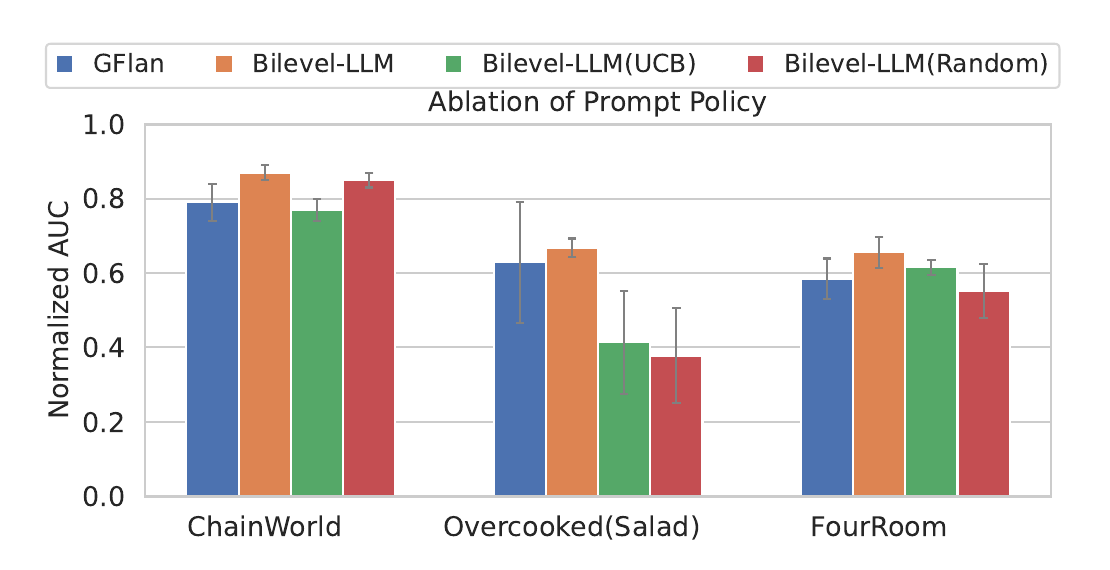}
    		}
    \subfigure[Ablation on multimodal]{
    \label{fig:symbolic}    		\includegraphics[width=3.5cm ,height=2.0cm]{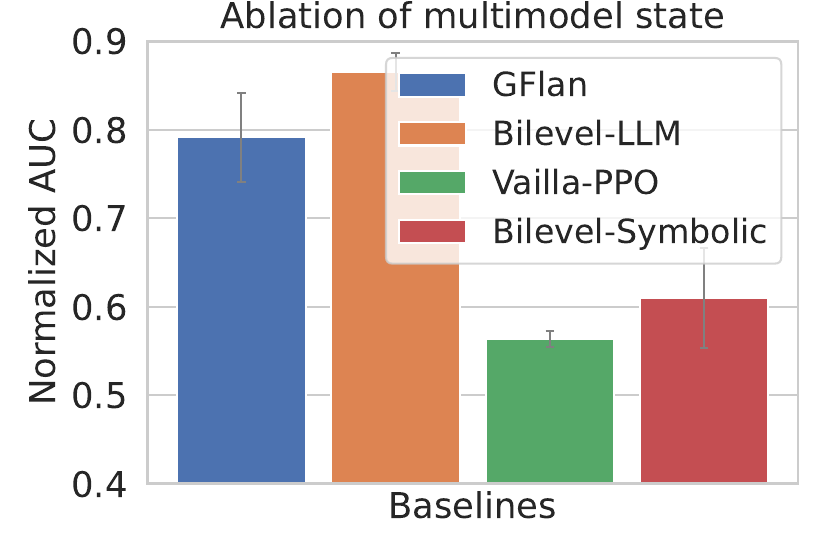}}
        \subfigure[{Ablation on auto prompt}]{
    \label{fig:autoprompt}    		

      \includegraphics[width=3.5cm ,height=2.0cm]
      {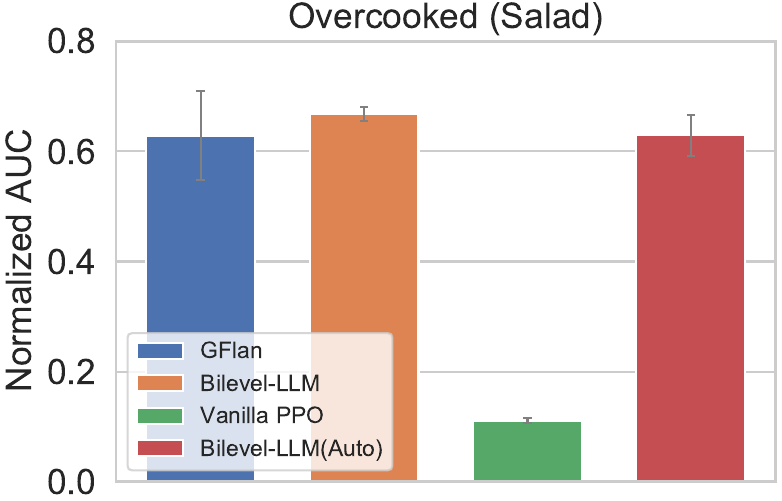}

\includegraphics[width=3.5cm ,height=2.0cm]
    {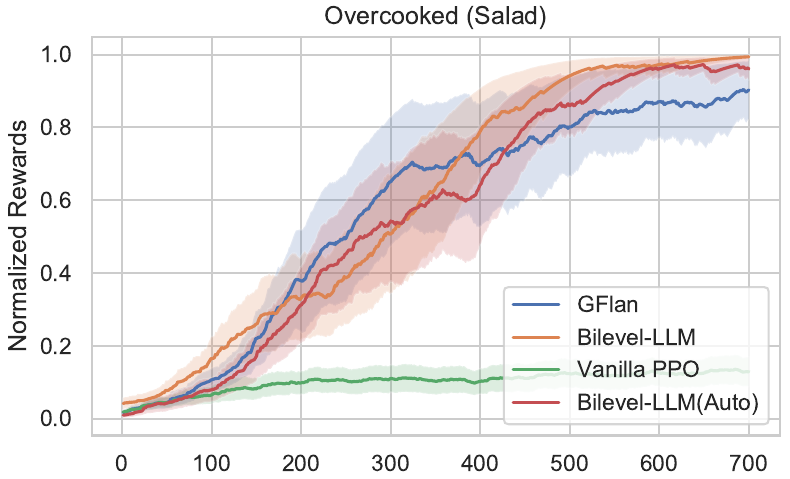}
}
        \vspace{-1.em}

    \caption{ Ablation studies. (a) The effect of different prompt generation strategies. (b) Verficiation of the effectiveness of \ourmethod~ under multimodal state representations on ChainWorld. (c)Automatically generate prompts on Overcooked(Salad). \textit{Left:} Normalized AUC reward. \textit{Right:}  Rewards during training.
    \vspace{-1.5em}
    }
\label{fig:ablation1}
\end{figure*}

\noindent\textbf{Comparison with baselines.} The results of comparisons with baselines are shown in \fig \ref{fig:baselines}, where we plot the normalized reward averaged over the whole training process. The training curves on all environments can be found in \fig \ref{fig:trainingcurvebaselines}. 
$\ourmethod$ outperforms other baselines in all environments also and exhibits a smaller standard error than the suboptimal GFlan. This indicates that $\ourmethod$ incorporating CoT reasoning can improve the task solving performance and convergence rate. In addition, GFlan outperforms Vanilla PPO on all environment especially on Overcooked. This suggests that using the pretrained LLM as an action policy is beneficial to decision-making due to the rich prior knowledge it contains and its stronger world rule summary and reasoning abilities. 
\newline
Furthermore, for most environments, GPT-3.5, GPT-3.5 (CoT Prompt), and ReAct struggle to solve decision-making tasks effectively. This indicates that although GPT-3.5 is powerful in generating useful high-level task solutions (thoughts), it faces challenges in long-term decision-making processes due to the complexity of the world model and rules in the environment. For instance, in the Tower of Hanoi game, GPT-3.5 is capable of identifying valid and invalid moves but struggles to reason through the correct move sequence from the start to the goal state, as illustrated in the Appendix \ref{appendix:tower_GPT-3.5}. 
Additionally, the task-solving ability of these methods heavily relies on the quality of human-designed prompts and CoT output process functions, thus requiring extensive human labor and limiting their generalisation ability. However, our method automatically selects a prompt per state and leverages reinforced action policy to interpret CoT outputs and collect experience, achieving better performance and generalising well. 

\noindent{\textbf{Does $\ourmethod$ learn to automatically generate prompts?} For most environments, we utilize the prompt candidates set generated by GPT-3.5 with the task description given. For Overcooked and Fourroom, there are obvious subtask divisions, thus we directly use these subtasks as prompt questions. 
{We also verify the automatically generated prompt candidates on Overcooked. \textit{$\ourmethod$-Auto} is displays the performance of \ourmethod~ when the prompt candidates are automatically generated by GPT-3.5}. As shown in \fig \ref{fig:autoprompt}, \textit{Bilevel-LLM-Auto} achieve similar rewards compared to $\ourmethod$ with human-designed prompt candidates. In addition, $\ourmethod$ and \textit{Bilevel-LLM-Auto} both outperform GFlan and exhibit lower variance. Specifically, after training the same number of episodes, $\ourmethod$ reaches a normalized reward around $1.0$ but GFlan only reaches around $0.9$. This suggests that CoT thoughts induced by appropriately selected prompts are helpful in solving complex decision-making tasks. In addition, our plug-and-play framework is able to utilise the automatically generated task-relevant prompts from GPT-3.5. 
}
Examples of automatically generated prompts can be found in Section \ref{autopromptappendix} of the Appendix.  

    


\subsection{Ablation Studies}
We conducted a series of ablation studies to confirm the usefulness of the components of \ourmethod~. In the following, we modified components of \ourmethod~ in order to validate the following claims: 

\noindent\textbf{Does the prompt policy with policy gradient improve performance?} In order to validate the claim that the prompts generated by \ourmethod~ lead to improved performance, we tested \ourmethod~ against the baseline \textit{$\ourmethod$ (Random)}, which is \ourmethod~ but with the prompt policy replaced so that we randomly select a prompt from the candidate set at each time step. In addition, \textit{$\ourmethod$ (UCB)} views the prompt selection from a candidate set as the multi-armed bandit problem and uses Upper Confidence Bound (UCB) to select the prompt. In this setting, the UCB algorithm does not consider the historical observation but only relies on the environment rewards, i.e., the minus entropy of the action policy to select a prompt. In addition, the UCB counts are reset for each episode. 
As shown in \fig \ref{fig:ablationprompt}, $\ourmethod$ outperforms all other prompt policy versions on all environments. 
The bad performance of \textit{$\ourmethod$ (UCB)} might be due to the lack of consideration of environmental state when performing prompt selection.
\begin{figure}
    
  \centering
{
    \includegraphics[width=3.5cm ,height=2.4cm]
      {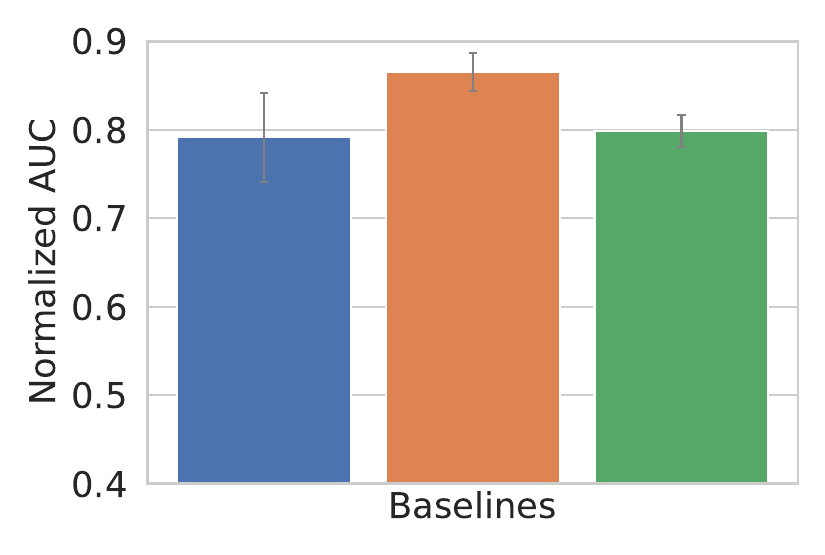}}
  {
    \includegraphics[width=3.5cm ,height=2.4cm]
    {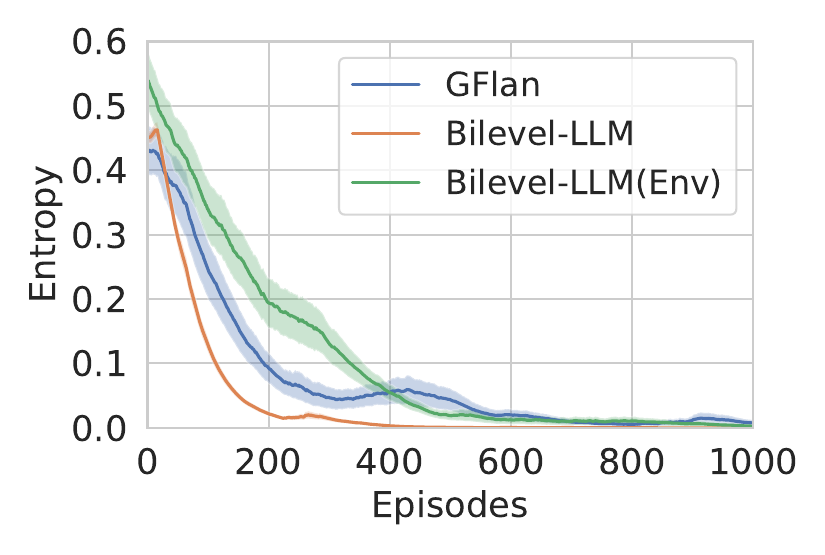}
}  
  \caption{{Ablation of the entropy objective on Chainworld (Partial). \textit{Left:} Normalized AUC reward. \textit{Right:} Entropy of the action policy. }}
    \vspace{-.5cm}

	\label{ablation2: env}
\end{figure}

\noindent\textbf{Does the entropy objective improve performance?} To validate the claim that the entropy objective leads to better performance we tested \ourmethod~ against the baseline 
\textit{$\ourmethod$ (Env)}, which replaces the negative entropy with the reward from the environment. As shown in Fig. \ref{ablation2: env}, $\ourmethod$ with entropy objective outperforms \textit{$\ourmethod$ (Env)} and exhibits lower entropy of the aciton policy.  

\noindent\textbf{Can the $\ourmethod$ framework accommodate multimodal state representation?} {We design a baseline \textit{$\ourmethod$-Symbolic}, where the action policy is replaced by that of Vanilla PPO, but taking both the embedding of the CoT output and symbolic environment observations as the input.} As shown in \fig \ref{fig:symbolic}, $\ourmethod$ outperforms GFlan and \textit{$\ourmethod$-Symbolic} outperforms Vanilla-PPO, which indicates that the utilization of prompt questions and CoT reasoning is helpful to improve the capability of action policies with both textual and symbolic state representation.

\section{Related Work}
\textbf{Reasoning with LLMs.}
Previous studies have confirmed that stage-by-stage reasoning significantly enhances the capability of LLMs to solve complex tasks such as mathematical and logistic reasoning problems. CoT\citep{wei2022chain} prompts containing a series of intermediate reasoning steps is shown to improve the inference ability of LLMs. Self-consistency \citep{wang2022self} marginalizes over several independent CoT results and then selects the most consistent answer. PAL \citep{gao2023pal} integrates executable programs into the CoT reasoning, addressing computation-related problems.
Besides using the prior world knowledge contained in LLMs, {ReAct \citep{yao2022react}, Tree-of-Thought (ToT) \citep{yao2023tree} and RAP \citep{hao2023reasoning} make use of from external environments or internal LLMs to produce reasoning traces.} 
ToT \citep{yao2023tree} and RAP  \citep{hao2023reasoning} explore extensively compared to CoT, which encompass multiple reasoning paths and construct a reasoning tree to determine the next crucial action. 
In this work, LLMs are applied to address multi-step decision-making problems, such as the Overcooked, where human-like solution reasoning is essential for taking actions.

\noindent\textbf{LLMs for RL.}
Due to the impressive reasoning capabilities of humans and the wealth of knowledge preceding LLMs, a series of studies have attempted to incorporate LLMs into planning algorithms to address decision-making tasks. ICPI \citep{brooks2022context} solves a number of simple interactive RL tasks (such as Maze) without the need for expert demonstrations or gradient computations, which uses LLMs as the world model and the rollout policy with historical interactions as in context examples. \citet{chen2023introspective} leverage historical trajectories to prompt LLM to generate the next step actions on the TextWorld game. 
GFlan\citep{carta2023grounding} aims to ground the LLM Flan-T5 \citep{rae2021scaling} on solving a textual interactive task named BabyAI-Text. 
In this approach, Flan-T5 serves as the action policy and is fine-tuned via PPO \citep{schulman2017proximal}.
 LFG \citep{shah2023navigation} utilises an LLM with a polling strategy to recommend and subsequently rank subgoals.
In our work, we integrate complex CoT reasoning into RL to enhance the quality of actions while eliminating the need for meticulous engineering to interpret LLM outputs.

\noindent\textbf{Entropy in RL.}
Entropy has been used extensively in RL as a tool for regularisation~\citep{mnih2016asynchronous,marl-book}. The policy in actor-critic methods is often trained with an additional term that aims to maximise the entropy of the learned actions, with the goal of exploring the environment without having a policy collapse early to suboptimal actions~\citep[e.g.][]{mnih2016asynchronous}.
A more formal use of entropy is explored in maximum entropy reinforcement learning~\citep{haarnoja2018soft, eysenbach2021maximum}, where the optimisation objective aims to learn the optimal policy that has the maximum entropy.
In this work, we take a different approach, and look at finding prompts that minimise the entropy of the action policy. Intuitively, this would push the CoT process to provide reasoning that makes the policy sure about its action. Such minimization of the entropy has also been explored: \citet{zhang2021hierarchical} formulate a hierarchical approach to intrinsic options, where entropy is minimised to improve the option sub-trajectories, and \citet{allahverdyan2018adaptive} consider entropy for decision making in the exploration-exploitation trade-off.

\noindent\textbf{Automated Prompt Engineering.}
The quality of prompts plays a crucial role in determining the output quality of LLMs. {Many works hand-craft desirable prompts such as the Generative Agents \citep{park2023generative} and ProAgent \citep{zhang2023proagent}. Apart from completely using human-crafted prompts, there are other studies that adopt different degrees of automation when generating meaningful prompts. For example, APE \citep{zhou2022large} and DLN \citep{sordoni2023deep} generate prompts from multiple examples and utilise LLM to rank the prompt candidates. PromptPG \citep{lu2022dynamic} trained an additional prompt selection network using the policy gradient, where the network generates the probability distribution over a predefined set of prompt examples. 
{We also aim to minimise human-labor on prompt engineering, we therefore preset a group of prompts and let the prompt policy choose depending on the environment observations.}}
\vspace{-0.05cm}
\section{Conclusion}
We introduce \ourmethod, a bilevel framework that is capable of learning introspective questions (in the form of prompts), then performing complex reasoning for guiding actions executed by an action-policy. 
The bilevel nature of the framework enables the accommodation of separate objectives for the two learning components, namely the prompt-generation policy uses an action-policy entropy minimisation objective which enables it to induce unambiguous and useful prompts to be fed to the action-policy. Meanwhile, the action-policy learns how to perform actions in the environment while making use of the CoT thoughts which it learns to interpret. We showed that this leads to a powerful framework that outperforms leading baselines in complex benchmark environments. 
We believe our framework takes an important step towards generalist artificial intelligence that is capable of introspection and complex decision-making. 
In this work, we only explore our framework on solving single-agent decision-making tasks, but neglect the prevalent multi-agent setting. Further work will extend our framework to encompass decision-making in multi-agent scenarios, possibly exploring the potential of leveraging the reasoning abilities of LLMs to uncover cooperation patterns or to model the behaviors of opponents.

\nocite{langley00}
\section*{Impact Statements} 
This paper introduces an advanced decision-making solver that combines the reasoning ability of LLMs with the online experience collection ability of RL algorithms. To enhance the CoT reasoning reliability of LLMs, a prompt policy is designed to select appropriate prompt questions from a candidate set based on the current environment state. Experimental results on popular game scenarios such as cooking and maze games, validate the superiority of our framework compared to powerful decision-making baselines like Vanilla PPO and GPT-3.5 with prompt strategies. 
Furthermore, this work has the potential to address complex real-world decision-making scenarios, such as autonomous driving and robotics. We believe that our work takes a significant step towards automatically solving complex decision-making tasks by combining powerful general tools with task-specific learning processes. 

\bibliography{ref}
\bibliographystyle{icml2024}
\appendix

\onecolumn 
\icmltitle{ Supplementary Material for Ask more, know better: Reinforce-Learned Prompt Questions for Decision Making with  Large Language Model
}

\section{Additional description about the methodology}
\label{'appendix:gradient'}
The following proof is a detailed derivation of the approximated gradient shown in \eqref{eq:pg_gradient} following the policy gradient method:
\begin{equation}
\begin{aligned}
    \nabla_\phi {J}(y|\pi_\theta,\pi_\phi,\pi^{re})
    &=\sum_{\tau\sim \pi_\theta,\pi_\phi,\pi^{re}} \nabla_\phi\rho(\tau) R_o(\tau)\\
    &=\sum_{\tau\sim \pi_\theta,\pi_\phi,\pi^{re}} \rho(\tau)\nabla_\phi\log \rho(\tau) R_o(\tau)\\
    &=\sum_{\tau\sim \pi_\theta,\pi_\phi,\pi^{re}} \rho(\tau)\sum_{t\geq0} \nabla_\phi\log\pi_\phi(p_t|o_t,\ldots,o_{t-j\wedge 0}) R_o(\tau)\\
     &=\mathbb{E}_{\tau\sim \pi_\rho,\pi_\theta,\pi^{re}}\sum_{t\geq0} \nabla_\phi\log\pi_\phi(p_t|o_t,\ldots,o_{t-j\wedge 0}) R_o(\tau)\\
    &\approx \frac{1}{N} \sum_{t\geq0} \nabla_\phi\log\pi_\phi(p_t|o_t,\ldots,o_{t-j\wedge 0}) R_o(\tau)\\
    &\approx \frac{1}{N} \sum_{t\geq0} \nabla_\phi\log\pi_\phi(p_t|o_t,\ldots,o_{t-j\wedge 0}) \hat{R}^o_t(\tau),
\end{aligned}
\label{eq:pg_gradient_dev}
\end{equation}
where $\hat{R}^o_t(\tau):= -\sum_{i\geq t}\left[\gamma^{i-t}\mathcal{H}^\pi_\theta(y_i)|y_i=(o_i,\upsilon_{i}), \upsilon_{i}\sim\pi^{\rm re}(i_t)\right]$ denotes the return-to-go from step $t$ to the end for the outer loop.
\begin{figure*}[h!]
 \vspace{-1em}
	\centering
   		\includegraphics[width=0.85 \linewidth]{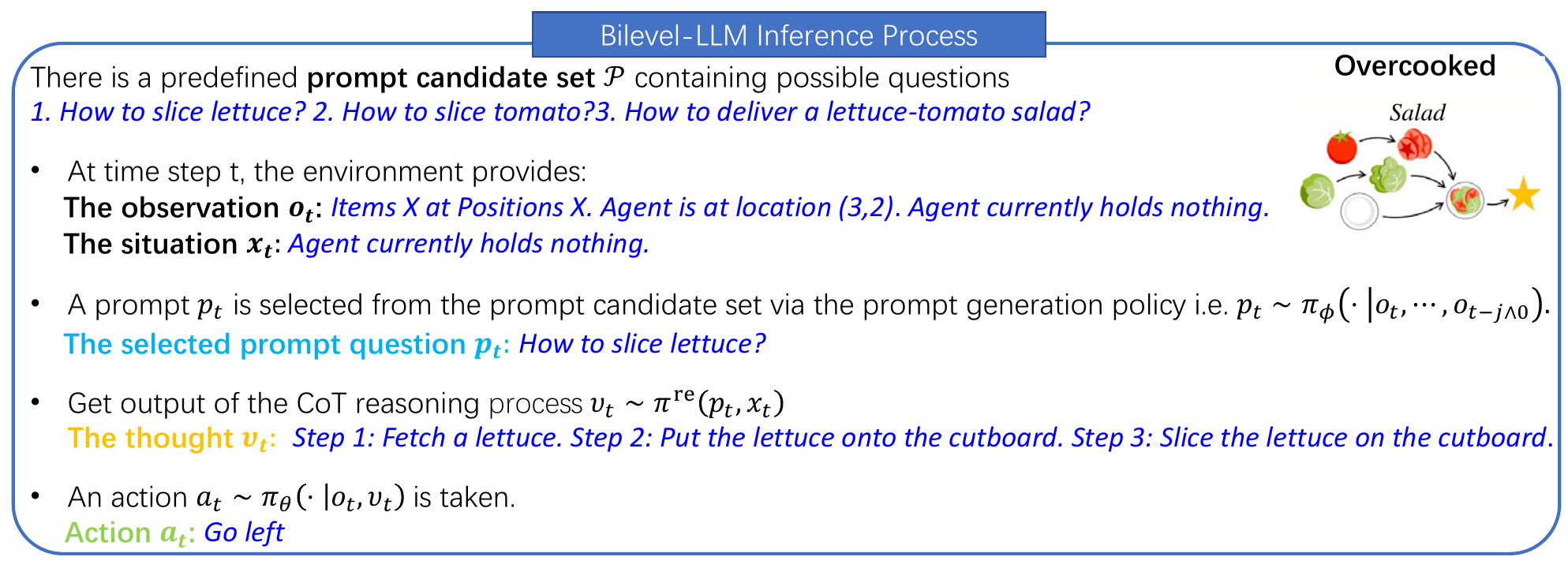}
    \caption{{An example of the step-by-step inference process of $\ourmethod$ on the Overcooked task.} 
    }
    \label{fig:illustrationprocess}
\end{figure*}
\section{Hyperparameter}
As shown in Table \ref{tab:parahanoi},\ref{tab:paralake},\ref{tab:parachainworld},\ref{tab:parafour},\ref{tab:paraover} we report the important hyperparameters for learnable baselines $\ourmethod$, GFlan and Vanilla PPO on all environments. Baselines GFlan and Vanilla PPO are trained by the PPO algorithm. For $\ourmethod$, the prompt generation and action policies are alternately trained with the other policy frozen. The parameter update frequency indicates the number of collected episodes for each update. 
We do a grid search to select hyperparameters. We tune the learning rate in the range of $[1e-3, 1e-4,1e-5]$, the batch size in the range of $[16, 32, 64,128]$, and the update frequency in the range of $[8,12,16,20,32,64]$. 

\newpage
\begin{table}[]
    \caption{The hyperparameters on Tower of Hanoi}

    \centering
    \begin{tabular}{c|cccc}
    \hline
    \multicolumn{5}{c}{\textbf{Action policy}}\\
    \hline
    \hline
    Baselines & Learning Rate & Epochs & Batch Size & Update Frequency\\
    \hline
        $\ourmethod$ & 1e-4& 4&64&64  \\
        GFlan & 1e-4& 4&128&64  \\
        Vanilla PPO  & 1e-3& 4&128&64 \\
    \hline
    \multicolumn{5}{c}{\textbf{Prompt generation policy}}\\
        \hline
    \hline

    Baselines & Learning Rate & Epochs & Batch Size & Update Frequency \\
        \hline
        $\ourmethod$ &1e-3  &4 &64 &64 \\
        GFlan &/ &/ &/ &/ \\
        Vanilla PPO & /&/&/&/\\
            \hline
        \end{tabular}
    \label{tab:parahanoi}

\end{table}

\begin{table}[]
    \caption{The hyperparameters on Frozen Lake}

    \centering
    \begin{tabular}{c|cccc}
    \hline
    \multicolumn{5}{c}{\textbf{Action policy}}\\
    \hline
    \hline
    Baselines & Learning Rate & Epochs & Batch Size & Update Frequency\\
    \hline
        $\ourmethod$ & 1e-5& 4&128&64  \\
        GFlan & 1e-5& 4&128&64  \\
        Vanilla PPO  & 1e-3& 4&128&64 \\
    \hline
    \multicolumn{5}{c}{\textbf{Prompt generation policy}}\\
        \hline
    \hline

    Baselines & Learning Rate & Epochs & Batch Size & Update Frequency \\
        \hline
        $\ourmethod$ &1e-5  &4 &128 &64 \\
        GFlan &/ &/ &/ &/ \\
        Vanilla PPO & /&/&/&/\\
            \hline
        \end{tabular}
    \label{tab:paralake}

\end{table}
\color{black}
\begin{table}[t!]
    \caption{The hyperparameters on ChainWorld}

    \centering
    \begin{tabular}{c|cccc}
    \hline
    \multicolumn{5}{c}{\textbf{Action policy}}\\
    \hline
    \hline
    Baselines & Learning Rate & Epochs & Batch Size & Update Frequency\\
    \hline
        $\ourmethod$ & 1e-4& 4&16&16  \\
        GFlan & 1e-4& 4&16&16  \\
        Vanilla PPO  & 1e-4& 4&16&16 \\
    \hline
    \multicolumn{5}{c}{\textbf{Prompt generation policy}}\\
        \hline
    \hline

    Baselines & Learning Rate & Epochs & Batch Size & Update Frequency \\
        \hline
        $\ourmethod$ &1e-4  &8 &32 &16 \\
        GFlan &/ &/ &/ &/ \\
        Vanilla PPO & /&/&/&/\\
            \hline
        \end{tabular}
    \label{tab:parachainworld}

\end{table}

\begin{table}[t!]
    \caption{The hyperparameters on FourRoom}

    \centering
    \begin{tabular}{c|cccc}
    \hline
    \multicolumn{5}{c}{\textbf{Action policy}}\\
    \hline
    \hline
    Baselines & Learning Rate & Epochs & Batch Size & Update Frequency\\
    \hline
        $\ourmethod$ & 1e-5& 4&32&16  \\
        GFlan & 1e-5& 4&32&16  \\
        Vanilla PPO  & 1e-5& 4&32&16 \\
    \hline
    \multicolumn{5}{c}{\textbf{Prompt generation policy}}\\
        \hline
    \hline

    Baselines & Learning Rate & Epochs & Batch Size & Update Frequency \\
        \hline
        $\ourmethod$ &1e-4  &8 &32 &16 \\
        GFlan &/ &/ &/ &/ \\
        Vanilla PPO & /&/&/&/\\
            \hline
        \end{tabular}
    \label{tab:parafour}

\end{table}

\begin{table}[t!]
    \caption{The hyperparameters on Overcooked}

    \centering
    \begin{tabular}{c|cccc}
    \hline
    \multicolumn{5}{c}{\textbf{Action policy}}\\
    \hline
    \hline
    Baselines & Learning Rate & Epochs & Batch Size & Update Frequency\\
    \hline
        $\ourmethod$ & 1e-4& 4&16&12  \\
        GFlan & 1e-4& 4&16&12  \\
        Vanilla PPO  & 1e-3& 4&16&12 \\
    \hline
    \multicolumn{5}{c}{\textbf{Prompt generation policy}}\\
        \hline
    \hline

    Baselines & Learning Rate & Epochs & Batch Size & Update Frequency \\
        \hline
        $\ourmethod$ &1e-3  &8 &64 &12 \\
        GFlan &/ &/ &/ &/ \\
        Vanilla PPO & /&/&/&/\\
            \hline
        \end{tabular}
    \label{tab:paraover}

\end{table}
\section{Additional Description of Experiments}
\subsection{Environments}
\label{envs}
\begin{figure*}[th!]
\centering
	\subfigure[Frozen Lake]{
 \label{fig:frozen}
    		\includegraphics[width=0.18 \linewidth]{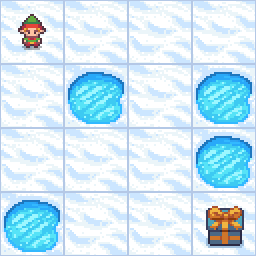}
    		}
	\subfigure[ChainWorld]{
 \label{fig:chainworld_env}
    		\includegraphics[width=0.18 \linewidth]{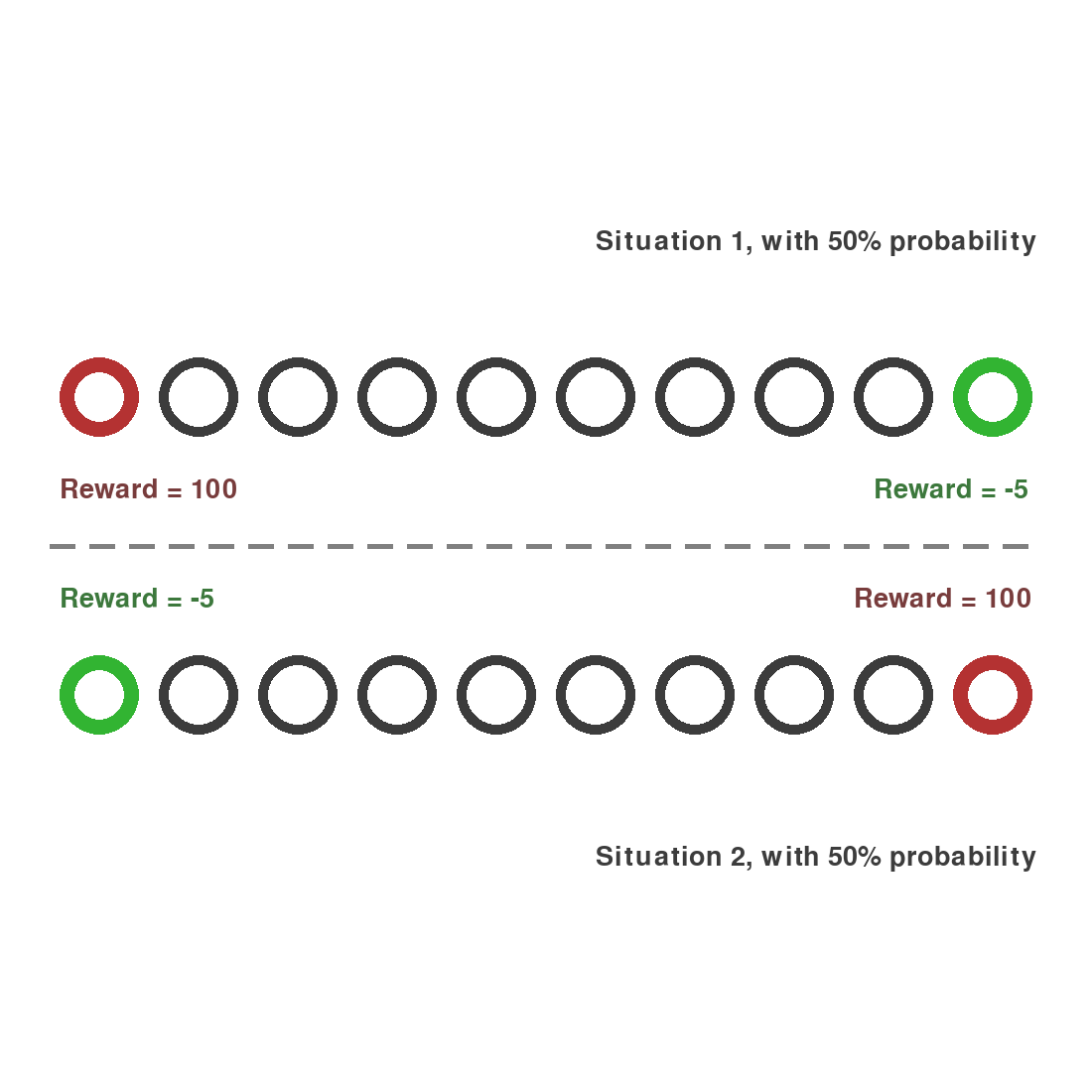}
    		}
      \vspace{-1.em}
    \subfigure[FourRoom]{
    \label{fig:fourroom_env}    		\includegraphics[width=0.18 \linewidth]{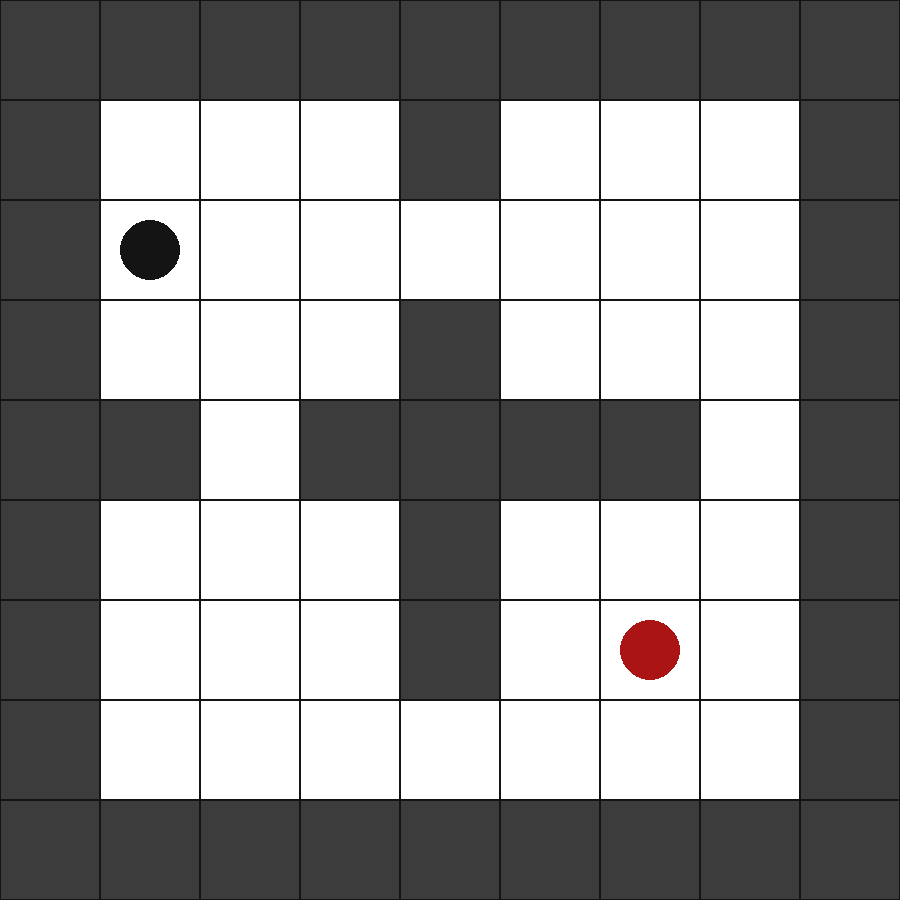}}
        \subfigure[Overcooked]{
    \label{fig:overcook_env}    		\includegraphics[width=0.18 \linewidth]{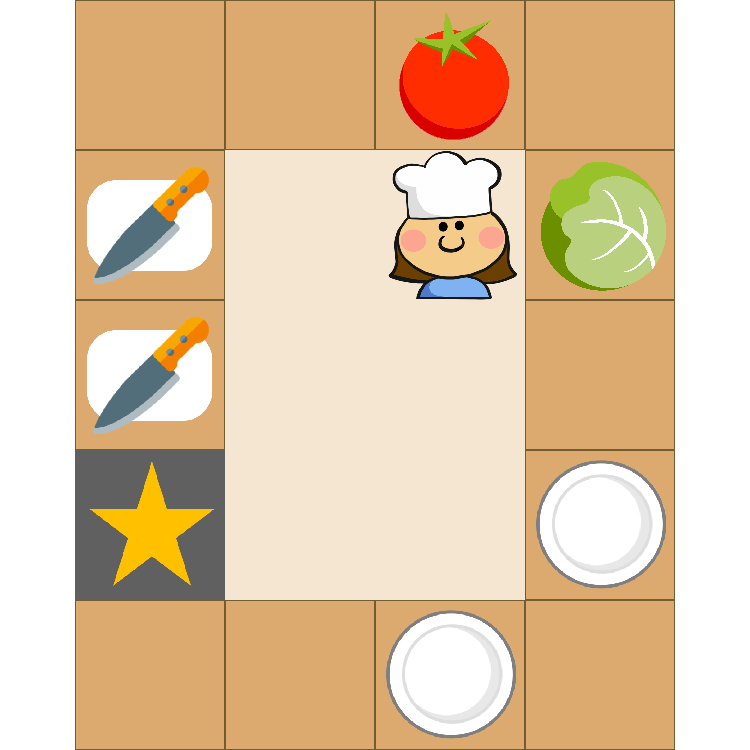}}
    \subfigure[Overcooked(Large)]{
    \label{fig:overcook_env_large}    		\includegraphics[width=0.18 \linewidth]{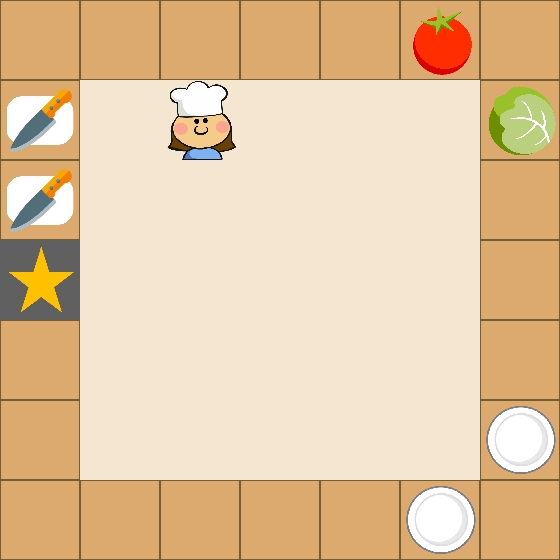}}
    \caption{{Environment Diagrams. (a) Illustration of the Frozen Lake environment. (b) Plots the ChainWorld Environment which has a 50\% probability of the goal in the left end of the chain and 50\% probability on the right. (c) The FourRoom Environment, which has four rooms and the position of the agent and the goal is randomly initialized. (d) The Overcooked Environment with a map size of $5\times4$. The agent needs to utilise the tools and material to make a salad and deliver it. (e) A large Overcooked layout with a map size of $7\times7$.}
    }
\label{fig:env_figs}
\end{figure*}
\textbf{Tower of Hanoi} \citep{hinz2013tower} This is a classic logic reasoning game that involves three disks of different sizes and three poles. The objective for the agent is to move the disks from an initial state to a goal state, adhering to the constraint that a larger disk cannot be placed on top of a smaller one. The agent gains reward $-1$ for invalid moves and reward $100$ for reaching the goal state. 
The textual observations are as the following structure:
\begin{itemize}
    \item \textit{Smallest disk on pole xxx, middle disk on pole xxx, largest on pole xxx.}
\end{itemize}
\textbf{Frozen Lake} This game involves crossing a frozen lake from start to goal positions without falling into any holes by walking over the frozen lake. The action policy has access only to the agent position due to the POMDP setting. We design auxiliary rewards for the training of all baselines. The agent reaching a place it has never been to gains a reward of +0.1, and reaching a place it has been to gains a reward of -0.01. Reaching the goal position gains a reward of 1.0. The textual observations are as the following structure:
\begin{itemize}
    \item \textit{The agent at position: $(x,y)$.}
\end{itemize}
\textbf{ChainWorld.}
The ChainWorld game contains a linear sequence of states and the available actions for the agent are go left or go right. The agent gains a reward $100$ at a random end of the chain and $-5$ at the other end, with $-1$ penalty for each move. At each episode, the award randomly appears on the left or right end, and the initial position of the agent is randomized, except for the ends. This game follows the POMDP setting, where only partial observation of the agent's position is available. Since the position with a reward of $100$ is randomized, the agent must learn to make decisions based on historical trajectory information. 
The textual observations are as the following structure:
\begin{itemize}
    \item \textit{You are at  \textit{\textless Position \textgreater}.}
\end{itemize}
\color{black}

\textbf{FourRoom.}
The FourRoom game is a navigation environment consisting of four rooms interconnected by four hallways in a circular arrangement. In this environment, the agent's initial position and the goal position are randomly set within these four rooms at the beginning of each game. 
 This environment provides a textual description instead of symbolic observations. The brief textual description consists of a list of template descriptions as following structure:
\begin{itemize}
    \item \textit{You are in \textit{\textless Room \textgreater}, goal is in  \textit{\textless Room \textgreater}. } 
    \item \textit{The left-handed hallway's position is \textit{\textless Position \textgreater}.}
    \item \textit{The right-handed hallway's position is \textit{\textless Position \textgreater}.}
    \item \textit{Your position is \textit{\textless Position \textgreater}. The goal's position is \textit{\textless Position \textgreater}.}
\end{itemize}
\color{black}
The \textit{\textless Room \textgreater} resembles one of the rooms, and the \textit{\textless Position \textgreater} is given as a list containing the x-coordinate and the y-coordinate. At each time step, the agent receives textual observations and decides whether to move in one of the following directions: up, down, left, or right within the grid. When the agent chooses to move, it incurs a penalty of $-2$ if the movement leads to an invalid location, such as attempting to traverse a wall or moving out of bounds. The agent is rewarded with $50$ when it successfully reaches the goal, and receives a reward of $-0.4$ in other cases.
\newline \textbf{Overcooked.}  
The Overcooked environment is a grid game where agent can move in four directions and interact with items in the map by moving towards item's direction when standing next it it. The goal for each agent is to make the correct meal and deliver to the star location. In our experiments, we used a text-based variant where the state information such as the name and location of each item, the location of the agent itself and the item the agent currently holds, are included in text:
\begin{itemize}
    \item \textit{Currently in the kitchen there are the following items and their location:}
    \begin{itemize}
        \item \textit{Name: \textit{\textless Item Name \textgreater}, Location: \textit{\textless Item Location \textgreater};}
        \item \textit{Name: \textit{\textless Item Name \textgreater}, Location: \textit{\textless Item Location \textgreater};}
        \item ......
    \end{itemize}
    \item \textit{\textless Agent Name \textgreater is at location \textit{\textless Agent Location \textgreater} and currently holds \textit{\textless Item Name \textgreater}}
\end{itemize}
\color{black}
We use an incremental reward shaping scheme where we give small reward {+1} to agent for fetching a correct material, medium reward {+5} for successfully chopping the material,  large reward +10 for making a dish and putting it on a plat, finally a reward {+100} for delivering the dish.
\color{black}
The state space of Overcooked(Large) reaches  $9.8\times10^{21}$ and state space of common Overcooked layouts with the map size $5\times4$ reaches $8.3\times10^{12}$.\color{black} Take the Overcooked(Large) as an example. As shown in \fig \ref{fig:env_figs}, there are 21 positions on the kitchen counter where plates, tomatoes, lettuce, chopped tomatoes, chopped lettuce, a salad or nothing can be placed. Each cutboard can be in one of 5 states, which includes being empty, having one tomato, one lettuce, chopped tomato, chopped lettuce. The agent can stand at one of 25 grids and face one of four directions. Each player can hold either a plate, a tomato, a lettuce, chopped tomatoes, chopped lettuce, a completed salad or nothing in their hands. Therefore, the total number of possible states for this layout is: $7^{21} 5^2 C_{25}^1 C_4^1 C_7^1\approx 9.8\times10^{21} $.

\color{black}

\subsection{Prompts and CoT examples}
\label{autopromptappendix}


\subsubsection{For Tower of Hanoi from GPT-3.5}
\label{hanoiprompt}
\textbf{Task Description:}
Environment consists of three pegs and 3 disks of different sizes which can slide onto the pegs. The puzzle starts with all disks stacked on the first peg in ascending order, with the largest at the bottom and the smallest on top. The objective of the game is to move all the disks to the third peg. The only legal moves are those which take the top-most disk from one peg to another, with the restriction that a disk may never be placed upon a smaller disk.
'States: Tuple of len N (number of discs) - elements: pole indices in [0,1,2]

Actions: Character between a and f (see descriptions below)

a:(0,1) - top disk of pole 0 to top of pole 1.

b:(0,2) - top disk of pole 0 to top of pole 2.

c:(1,0) - top disk of pole 1 to top of pole 0.

d:(1,2) - top disk of pole 1 to top of pole 2.

e:(2,0) - top disk of pole 2 to top of pole 0.

f:(2,1) - top disk of pole 2 to top of pole 1.

The initial state (0,0,0) the three disks are all on pole 0.

The target state: (2,2,2) the three disks are all on the pole 2.

Please give 3 critical prompt questions to prompt subsequent cot that is helpful for solving this game. 

For example, one possible prompt question is "What are the valid moves for a given state?"

\textbf{Prompt Candidates} The prompt candidates given by GPT-3.5 are:
\begin{enumerate}
    \item What are the valid moves for a given state?

    \item What specific conditions must be satisfied to ensure a move is valid in the Tower of Hanoi game? Can you provide examples of both valid and invalid moves?

    \item Can you devise a plan or strategy to systematically move from the current state to the goal state?
\end{enumerate}

\textbf{CoT Examples:}
For example, at state (0, 0, 0), the smallest disk on pole 0, middle disk on pole 0, largest on pole 0. The CoT reasoning results corresponding to above $3$ question are:
\begin{enumerate}
    \item The valid moves for the current state (0, 0, 0) are:
    
- a: Move the smallest disk from pole 0 to pole 1

- b: Move the smallest disk from pole 0 to pole 2

    \item Valid moves in the Tower of Hanoi game must satisfy the condition that a disk can only be placed onto a larger disk. For example, moving the top disk from pole 0 to pole 1 would be a valid move as long as the top disk on pole 1 is larger than the one being moved. An example of an invalid move at the current state would be trying to move the middle disk from pole 0 to pole 1, as the middle disk is larger than the top disk on pole 1.

    \item Start by moving the smallest disk to pole 2. Then move the middle disk to pole 1. Finally, move the largest disk to pole 2. This will solve the puzzle.

\end{enumerate}
\subsubsection{For Frozen Lake From GPT-3.5}
\textbf{Task Description:}
 Frozen lake involves crossing a frozen lake from Start(S) to Goal(G) without falling into any Holes(H) by walking over the Frozen(F) lake.  
 Reach goal(G): +1
Reach hole(H): 0
Reach frozen(F): 0

map is: 4x4:

    SFFF,
    
    FHFH,
    
    FFFH,
    
    HFFG
    
positions (x,y)

Start(left up) position: (0,0)

Goal(right down) position: (3,3)

actions: go left(x,y-1); go right(x,y+1); go up(x-1,y); go down (x+1,y)

Please provide $4$ prompt questions to subtasks of an agent from the start to goal position and with the layout information being taken into account?

\textbf{Prompt Candidates}
The prompt candidates given by GPT-3.5 are:
\begin{enumerate}
    \item What is the next action an agent should take from the current position in order to reach the Goal position (3,3) while avoiding any Holes on the Frozen lake?
    \item Considering the current layout of the Frozen lake, what are the possible actions an agent can take from its current position (x, y) to move closer to the Goal position?
    \item  How can an agent navigate the Frozen lake layout to reach the Goal position (3,3) by following a sequence of actions without falling into any Holes?
    \item Provide a series of actions an agent can take from the current position to the Goal position (3,3) that ensures a successful traversal over the Frozen lake without encountering any Holes.
\end{enumerate}

\textbf{CoT Examples:}
For example, at position (0, 0), the CoT reasoning results corresponding to above $4$ question are:
\begin{enumerate}
    \item The agent should consider going right (x, y+1) from the Start position (0,0) to step onto the Frozen tile (F) and advance towards the Goal position (3,3). This action avoids Holes and aligns with the overall goal of reaching the destination.

    \item Given the current layout, the agent, positioned at (0,0), can choose between going right (x, y+1) or down (x+1, y) to navigate the Frozen lake. Both options avoid Holes and contribute to the agent\'s progress toward the Goal (3,3).
    \item To successfully navigate the Frozen lake layout and reach the Goal (3,3), the agent, starting at (0,0), should prioritize moving right (x, y+1) to stay on the Frozen tiles (F) and away from Holes. Subsequent actions should be planned to maintain this trajectory.
    \item To ensure a successful traversal from Start (0,0) to Goal (3,3), the agent should follow the sequence: right (x, y+1), right (x, y+1), down (x+1, y), down (x+1, y), right (x, y+1), reaching the Goal while staying on Frozen tiles (F) and avoiding Holes.
\end{enumerate}
\subsubsection{For ChainWorld From GPT-3.5}
\label{chainworldprompt}
\textbf{Task Description:}
There is a chain world game. A chain of states in it, the agent can go left or go right. An agent can gain the reward -5 in the one side of the chain and gain reward 100 in the otherside of the chain. How to prompt the agent to move to gain high reward please give 3 simple prompts when left side with 100 reward or right side or unknown side with 100 reward respectively. Please give 3 simple prompts for solving this task.

\textbf{Prompt Candidates:} 
\begin{enumerate}
    \item Head left to discover a treasure trove of 100 points.
    \item Go right to seize the opportunity for a generous 100-point prize.
    \item Embrace the unknown, as 100 points could await in any direction.
\end{enumerate}

\subsubsection{For FourRoom from human design}
\textbf{Task Description:} In a four room game, there are four rooms(0,1,2,3) connected by four hallways. To move through different rooms, agent can only go through the hallways. The agent can choose a left-handed manner or a right-handed manner to move through different rooms. The agent's initial position can be either in a room or in a hallway, and the goal position can be in any room. The objective for the agent is to reach the goal.

\textbf{Prompt Candidates:}
\begin{itemize}
    \item You are in the same room as the goal, try to reach the goal.
    \item Goal is not in current Room; You should explore left-handed rooms to reach the goal.
    \item Goal is not in current Room; You should explore right-handed rooms to reach the goal.
    \item You are in the hallway between two Rooms. Go left-handed room, then explore.
    \item You are in the hallway between two Rooms. Go right-handed room, then explore.
\end{itemize}

\textbf{CoT Examples:}
\begin{enumerate}
    \item Order: Goal is not in current Room; You should explore left-handed rooms to reach the goal. \newline \newline Observations: You in Room1. Goal in Room3. Entrance to Left-handed hallway: [6, 4]. Entrance to Right-handed hallway: [4, 6]. Goal position: [1, 1]. \newline \newline CoT: "You in Room1, goal in Room3. Goal is not in current room. To move through different rooms, you can only go through hallways. Enter the left-handed hallway. How do you choose a step?"
    \item Order: You are in the hallway between two Rooms. Go right-handed room, then explore. \newline\newline Observations: You in hallway. Goal in Room3. Entrance to Left-handed hallway: [6, 4]. Entrance to Right-handed hallway: [4, 6]. Goal position: [1, 1]. \newline\newline CoT: "You in hallway. Goal is not in current hallway. Go to the right-handed room entrance. How do you choose a step?"
\end{enumerate}

\subsubsection{For Overcooked (Salad) from human design}
\textbf{Task Description:} 
I would like you to help me work with an AI agent called "agent1" in a kitchen environment similar to the video game Overcooked. Inside the kitchen there are the following items: ["tomato", "lettuce", "plate0", "plate1","cutboard0", "cutboard1"]. There are also the following functions that you can use to make agent1 take actions: \newline \newline agent1.fetch(item: str) - go to the item's location and pick it up (item will be in agent1's hand), only the cutboard cannot be picked up or fetched \newline agent1.put\_onto(item: str) - put the object agent1 has in hand onto the item \newline agent1.slice\_on(item: str) - slice food (item has to be "cutboard0") \newline agent1.deliver(None) - deliver the cooked food. \newline Note that the agent can only hold one item at a time.


\textbf{Prompt Candidates:}
\begin{enumerate}
    \item Question: How would you instruct "agent-1" to start making a sliced lettuce? You do not need to fetch the cutboard nor delivering the food. Think about what tasks need to be accomplished step by step. task\_queue = []
    \item Question: How would you instruct "agent-1" to start making a sliced tomato? You do not need to fetch the cutboard nor delivering the food. Think about what tasks need to be accomplished step by step. task\_queue = []
    {\item Question: How would you instruct "agent-1" to make and deliver a lettuce-tomato salad? Think about what tasks need to be accomplished step by step. task\_queue = []}
\end{enumerate}

\textbf{CoT Examples:} {With the sliced lettuce and the sliced tomato on the map and an empty hand, CoT reasoning outputs corresponding to above three questions are:}
\begin{enumerate}
    \item task\_queue = [] \newline \# Step 1: Fetch a lettuce \newline
    task\_queue.append((agent1.fetch, "lettuce")) \newline
    \# Step 2: Put the lettuce onto the cutboard (assuming cutboard0 is available)\newline
    task\_queue.append((agent1.put\_onto, "cutboard0")) \newline    
    \# Step 3: Slice the lettuce on the cutboard \newline task\_queue.append((agent1.slice\_on, "cutboard0"))
    \item  task\_queue = [] \newline \# Step 1: Fetch a tomato \newline
    task\_queue.append((agent1.fetch, "tomato")) \newline
    \# Step 2: Put the tomato onto the cutboard (assuming cutboard0 is available) \newline task\_queue.append((agent1.put\_onto, "cutboard0")) \newline 
    \# Step 3: Slice the tomato on the cutboard \newline task\_queue.append((agent1.slice\_on, "cutboard0"))
    \item task\_queue = [] \newline \# Step 1: Fetch a plate (choose either plate0 or plate1) \newline task\_queue.append((agent1.fetch, "plate0")) \newline 
    \# Step 2: Put the sliced lettuce onto the plate \newline task\_queue.append((agent1.put\_onto, "plate0")) \newline 
    \# Step 3: Fetch the sliced tomato \newline 
    task\_queue.append((agent1.fetch, "tomato")) \newline
    \# Step 4: Put the sliced tomato onto the plate \newline 
    task\_queue.append((agent1.put\_onto, "plate0")) \newline 
    \# Step 5: Deliver the lettuce-tomato salad \newline 
    task\_queue.append((agent1.deliver, None))
\end{enumerate}

\subsubsection{For Overcooked (Salad) From GPT-3.5}
\textbf{Task Description:} I would like you to help me work with an AI agent called "agent1" in a kitchen environment similar to the video game Overcooked. Inside the kitchen there are the following items: ["tomato", "lettuce", "plate0", "plate1","cutboard0", "cutboard1"]. There are also the following functions that you can use to make agent1 take actions: \newline \newline agent1.fetch(item: str) - go to the item's location and pick it up (item will be in agent1's hand), only the cutboard cannot be picked up or fetched \newline agent1.put\_onto(item: str) - put the object agent1 has in hand onto the item \newline agent1.slice\_on(item: str) - slice food (item has to be "cutboard0") \newline agent1.deliver(None) - deliver the cooked food. \newline Note that the agent can only hold one item at a time. Please provide three prompt questions about how to maximize reward.

\textbf{Prompt Candidates:} 
\begin{enumerate}
    \item Ingredient Preparation How can agent1 efficiently fetch the "tomato" and "lettuce" from their initial locations to start the salad preparation process? 
    \item Assembly Process Once agent1 has the "tomato" and "lettuce" in hand, how should it proceed with assembling the salad on the plates ("plate0" and "plate1")?
    \item Delivery Strategy What strategy should agent1 follow to deliver the prepared salads efficiently? 
\end{enumerate}

\textbf{CoT examples:} 

\begin{enumerate}
    \item Consider the order in which the ingredients need to be fetched and whether any preprocessing steps, such as slicing on the cutboard, are required.
    \item Explore a sequence of actions involving fetching, putting onto the plates, and any other necessary steps to assemble a complete salad on each plate. 
    \item  Consider factors such as the proximity of the plates to the delivery location and the ability of agent1 to carry only one item at a time. Think about optimal routes and whether any coordination is needed for efficient deliveries.
\end{enumerate}

\section{Additional Results}
\subsection{Training Performance} 
\begin{figure*}[th!]
	\centering
 \includegraphics[width=.5 \linewidth]{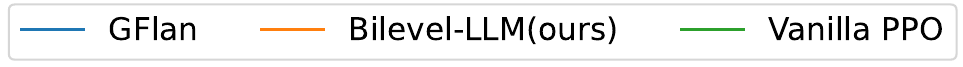}
    		
       {
   		\includegraphics[width=0.32 \linewidth]{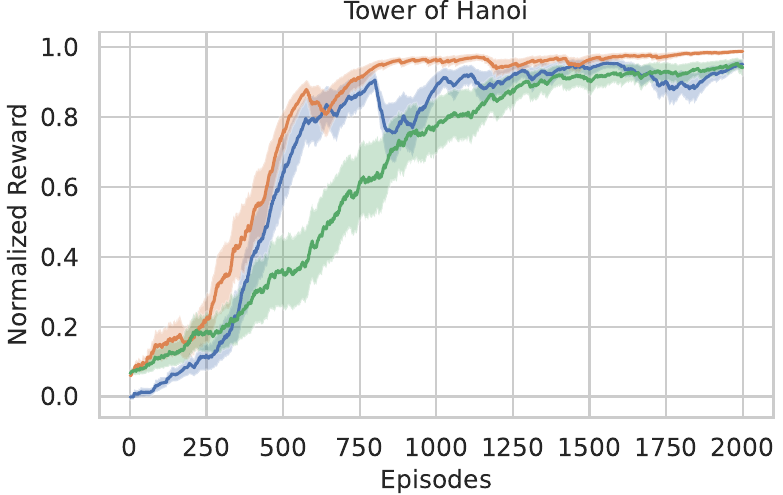}}
      {
   		\includegraphics[width=0.32 \linewidth]{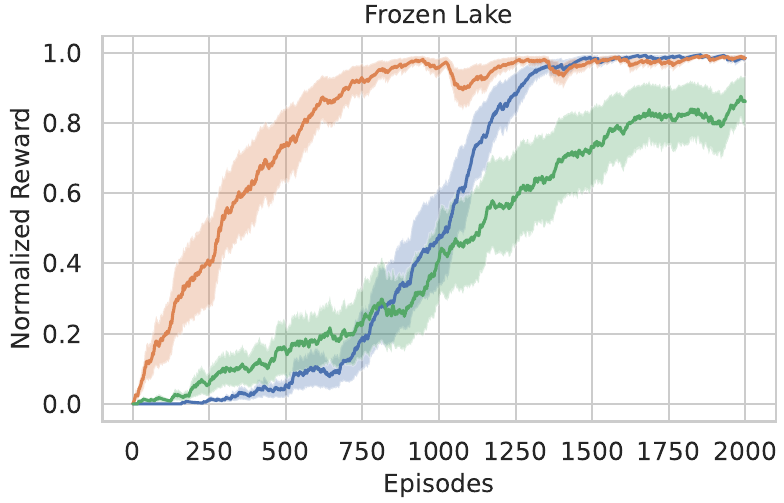}}
    {
   		\includegraphics[width=0.32 \linewidth]{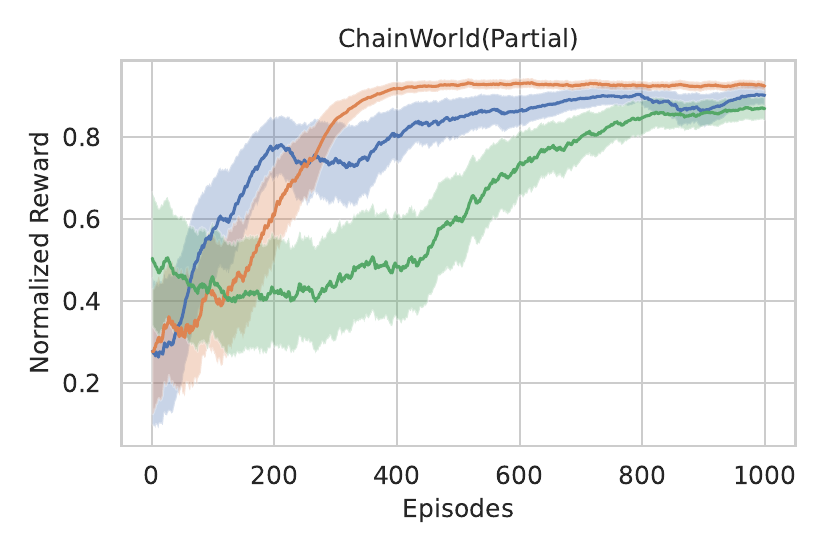}}
  {
     		\includegraphics[width=0.32 \linewidth]{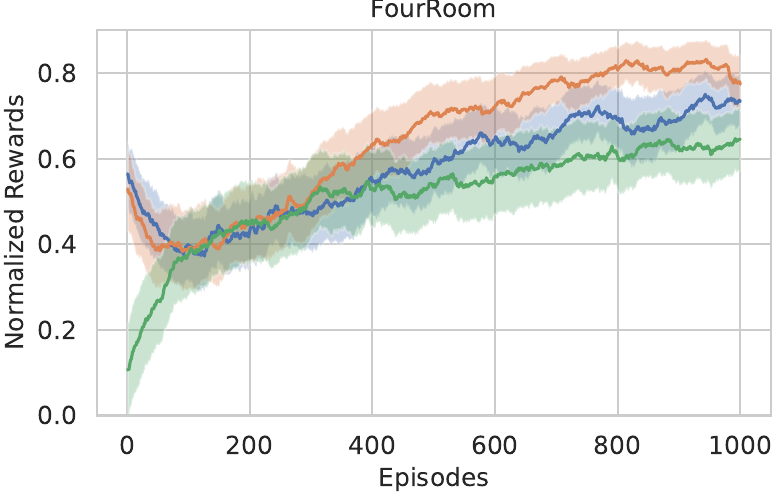}}
{
     		\includegraphics[width=0.32 \linewidth]{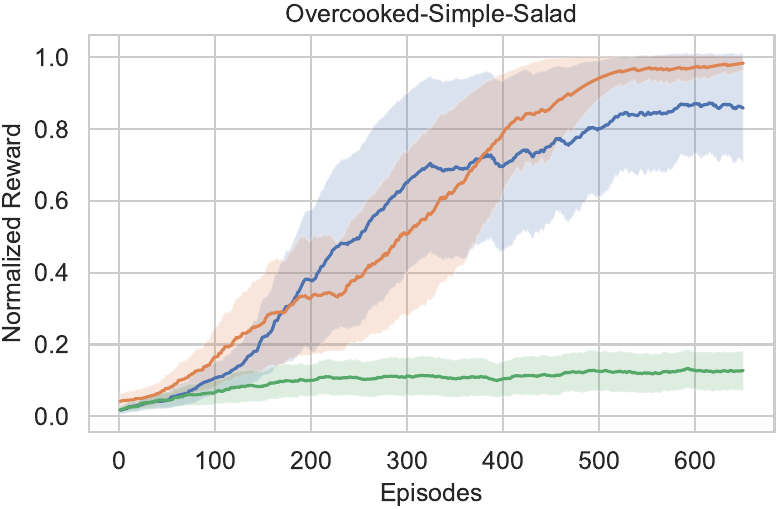}}
{
     		\includegraphics[width=0.32 \linewidth]{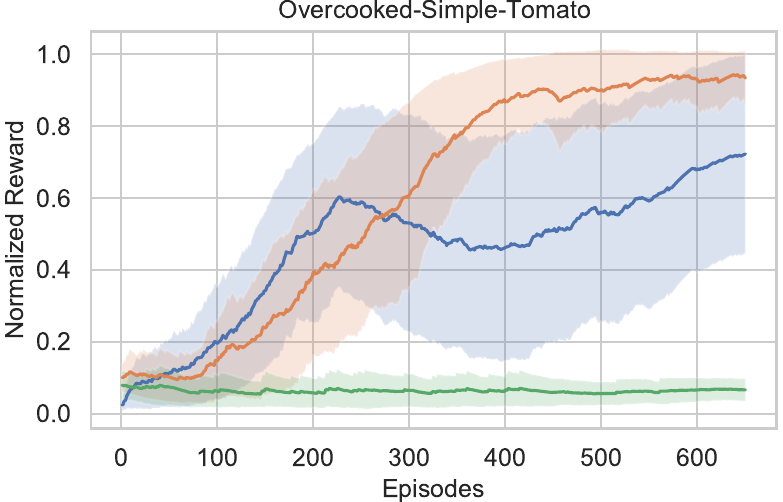}}
       {
     		\includegraphics[width=0.32 \linewidth]{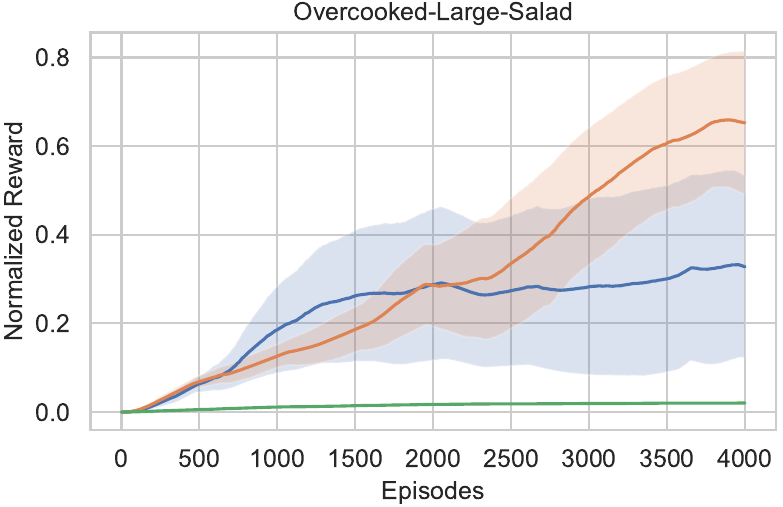}}

    \caption{ {Training curves of baselines. We plot the average and standard error of normalized rewards over $5$ seeds.}   
    }
\label{fig:trainingcurvebaselines}
\end{figure*}

{\textbf{Can the training of $\ourmethod$ framework converge?} Figure \ref{fig:trainingcurvebaselines} shows the training curves of the normalized rewards of the baselines ($\text{Our Method}$, GFlan, Vanilla PPO). These results indicate that, on these decision-making tasks, especially in the complex Overcooked scenario, $\text{Our Method}$ outperforms the other baselines. It converges faster and exhibits lower variance. Additionally, the training curves of our method appear smoother than that of the others, suggesting that our bilevel framework is more stable to converge. This may be attributed to the minimizing entropy objective for prompt generation policy and the utilisation of rich human knowledge assisting in decision-making on challenging scenarios.  }

\subsection{Analysis of GPT-3.5 reasoning ability for the Tower of Hanoi}
\label{appendix:tower_GPT-3.5}
The GPT-3.5 is able to reason the valid moves but fails to solve this pluzze game:

\paragraph{Identify valid moves}:
\newline 
\textit{Prompt question: "What are the valid moves for a given state?"
}\\
\textit{The valid moves for state (0, 0, 1) are:\\
- Move the smallest disk from pole 0 to pole 1 (a) \textcolor{blue}{Correct}\\ 
- Move the middle disk from pole 0 to pole 2 (b) \textcolor{blue}{Correct}\\
}
Thus the GPT-3.5 has the potential to identify valid moves, and the blue text is human annotation. 
\paragraph{Output the solution sequence from the current state to the goal state} The output of GPT-3.5 is: 
\\
\textit{
Based on the current state (1, 1, 2), you can follow these steps to systematically move towards the goal state (2, 2, 2):
\\
1. Move the smallest disk from pole 1 to pole 2. (1,2)\\ 
2. Move the middle-sized disk from pole 1 to pole 0. (1,0)\\
3. Move the smallest disk from pole 2 to pole 0. (2,0) \textcolor{blue}{Inappropriate moves, moving far from the goal.}\\ 
4. Move the smallest disk from pole 1 to pole 2. (1,2) \textcolor{blue}{incorrect move, the smallest disk currently on pole 2.}\\
5. Move the largest disk from pole 0 to pole 1. (0,1)\\
6. Move the smallest disk from pole 0 to pole 2. (0,2) \textcolor{blue}{incorrect move, the smallest disk currently on pole 2.}\\
7. Move the smallest disk from pole 1 to pole 2. (1,2) \textcolor{blue}{incorrect move, the smallest disk currently on pole 2.}\\}
Thus, the GPT-3.5 is struggle to solve this game, and the blue text is human annotation.
\subsection{Additional Ablation Study Results on ChainWorld}
\begin{figure*}[th!]
	\centering
 \includegraphics[width=1.0 \linewidth]{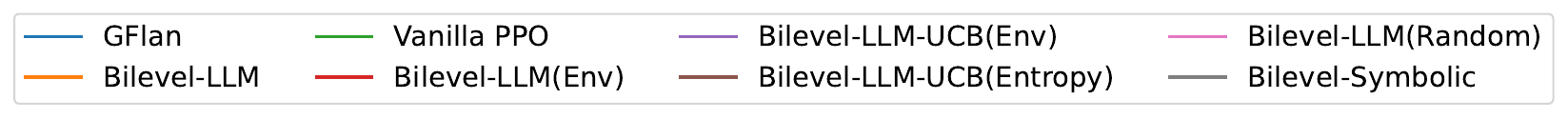}
    		
	\subfigure[Training Curve]{
 \label{fig:chaintraining}
    		\includegraphics[width=0.32 \linewidth]{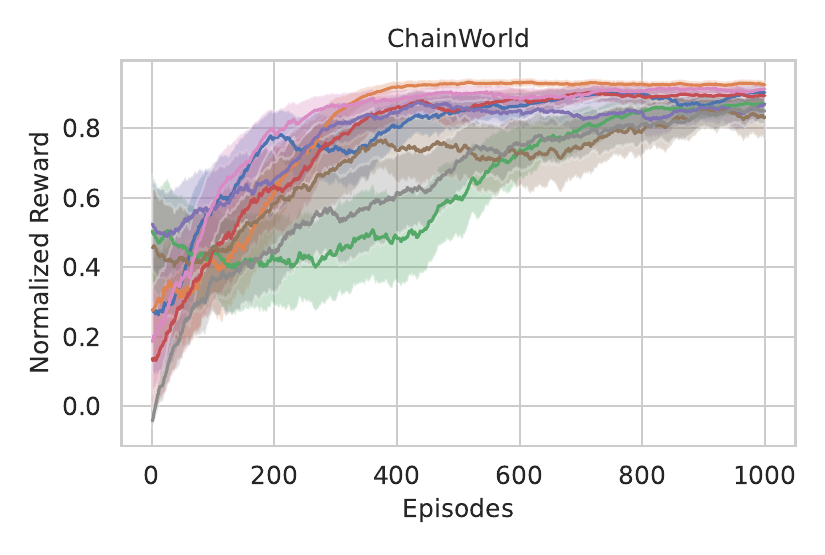}
    		}
      \vspace{-1.em}
    \subfigure[AUC Reward]{
    \label{fig:chainbar}    		\includegraphics[width=0.32 \linewidth]{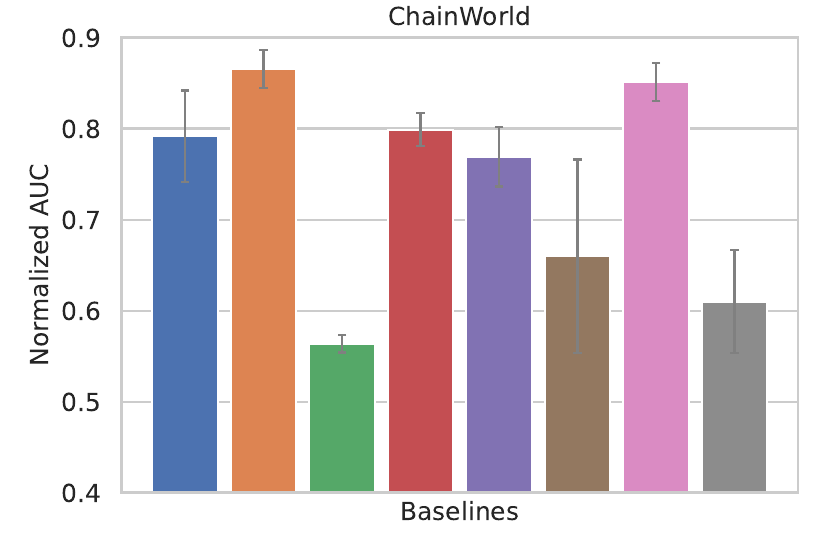}}
        \subfigure[Entropy]{
    \label{fig:chainentropy}    		\includegraphics[width=0.32 \linewidth]{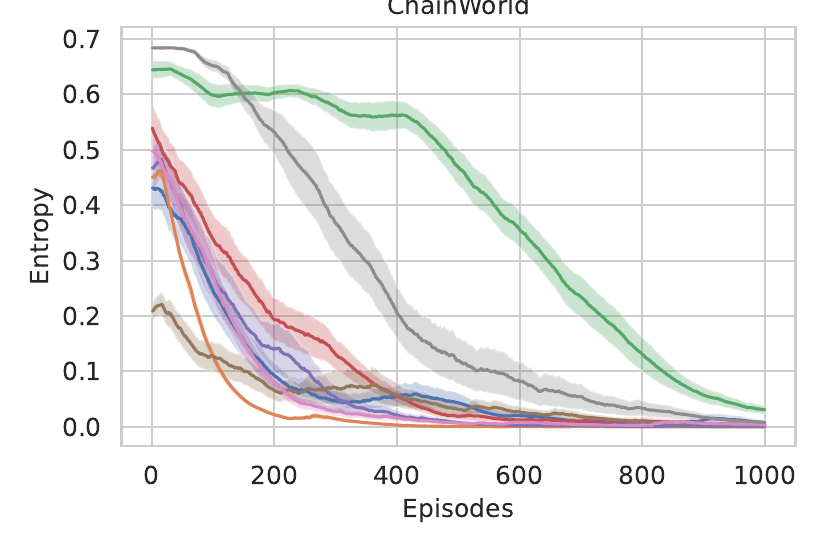}}
    \caption{ Results on Chainworld(Partial). (a) Training curves of baselines. We plot the average and standard error of rewards over $5$ seeds. (b) The average reward of baselines over all training episodes. (c) The entropy of action policy during training.   
    }
\label{fig:chainworld}
\end{figure*}

The performance of $\ourmethod$, compared to other baselines in the  ChainWorld environment is illustrated in \fig \ref{fig:chainworld}. Notably, as shown in \fig \ref{fig:chainentropy}, $\ourmethod$ exhibits the most rapid reduction in action policy entropy. This observation suggests that $\ourmethod$ shows the potiential to effectively learn optimal prompts while guiding the action policy to focus on the optimal policies through the bi-level optimization. Consequently, as depicted in \fig \ref{fig:chaintraining} and \fig \ref{fig:chainbar}, $\ourmethod$ outperforms other baselines at the end of training as well as the AUC reward which represents the average performance throughout the training.

\subsection{Additional Ablation Study Results on ChainWorld}

\begin{figure*}[th!]
	\centering
 \includegraphics[width=1.0 \linewidth]{fig/legendablation.pdf}
	\subfigure[Training Curve]{
 \label{fig:fourtraining}
    		\includegraphics[width=0.32 \linewidth]{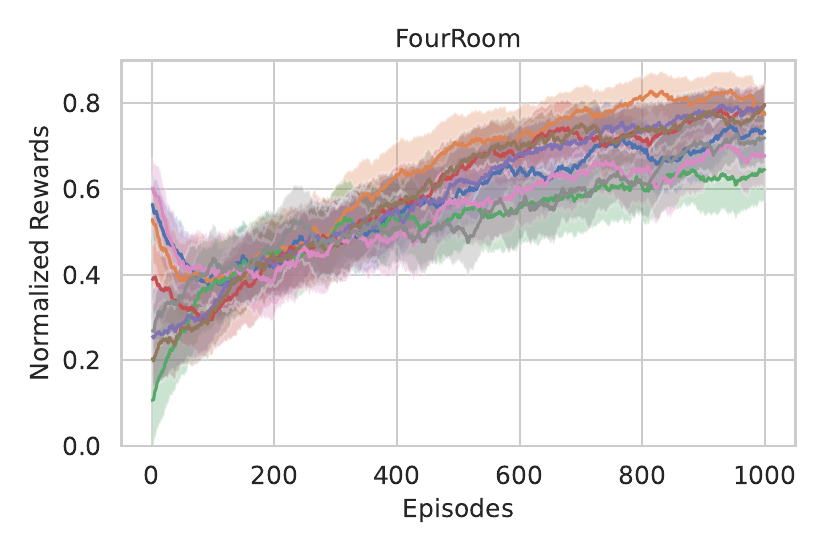}
    		}
      \vspace{-1.em}
    \subfigure[AUC Reward]{
    \label{fig:fourbar}    		\includegraphics[width=0.32 \linewidth]{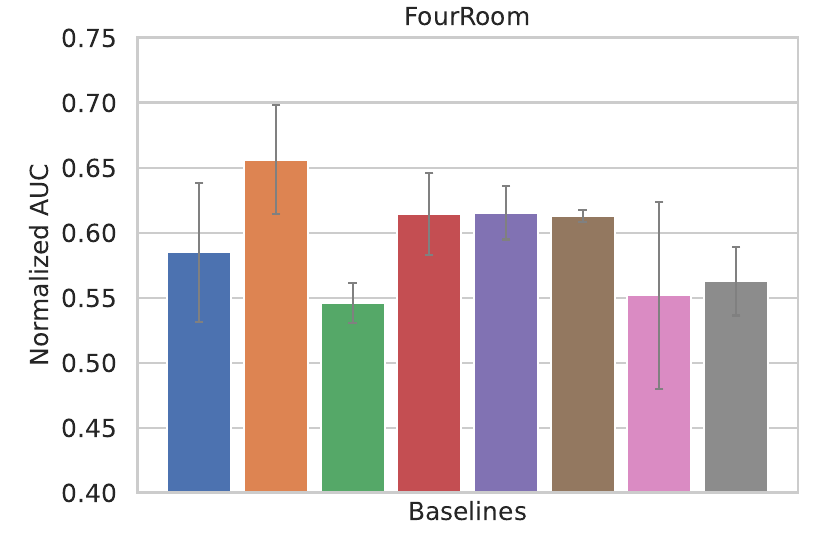}}
        \subfigure[Entropy]{
    \label{fig:fourentropy}    		\includegraphics[width=0.32 \linewidth]{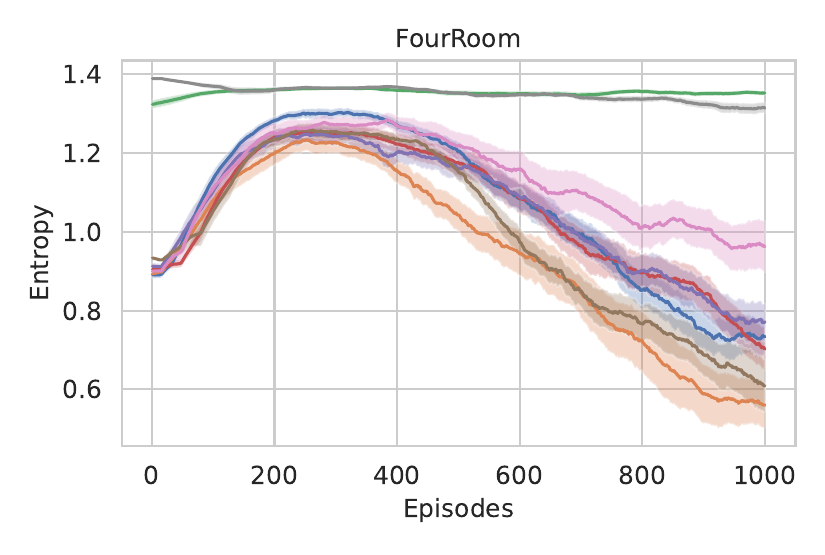}}

    \caption{ Results on FourRoom. (a) Training curves of baselines. We plot the average and standard error of rewards over $5$ seeds. (b) The average reward of baselines over all training episodes. (c) The entropy of action policy during training. 
    }
\label{fig:fourroom}
\end{figure*}

In the FourRoom environment, methods utilizing the LLM performs an increase of entropy in the first 200 episodes and turns to decay after 400 episodes. Examining \fig \ref{fig:fourtraining} and \fig \ref{fig:fourbar} concurrently, an interesting trend emerges:  the reduction in entropy often coincides with a decrease in reward. Conversely, an increase in entropy tends to correspond with a decline in performance, and the results supports the effectiveness of $\ourmethod$.

\subsection{Additional Ablation Study Results on ChainWorld}

\begin{figure*}[th!]
	\centering
 \includegraphics[width=1.0 \linewidth]{fig/legendablation.pdf}
	\subfigure[Training Curve]{
 \label{fig:overcooktraining}
    		\includegraphics[width=0.32 \linewidth]{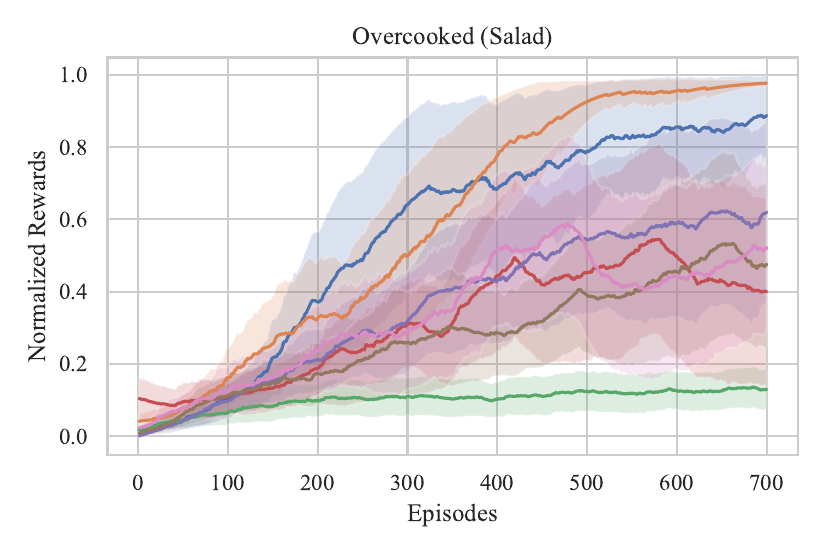}
    		}
      \vspace{-1.em}
    \subfigure[AUC Reward]{
    \label{fig:overcookbar}    		\includegraphics[width=0.32 \linewidth]{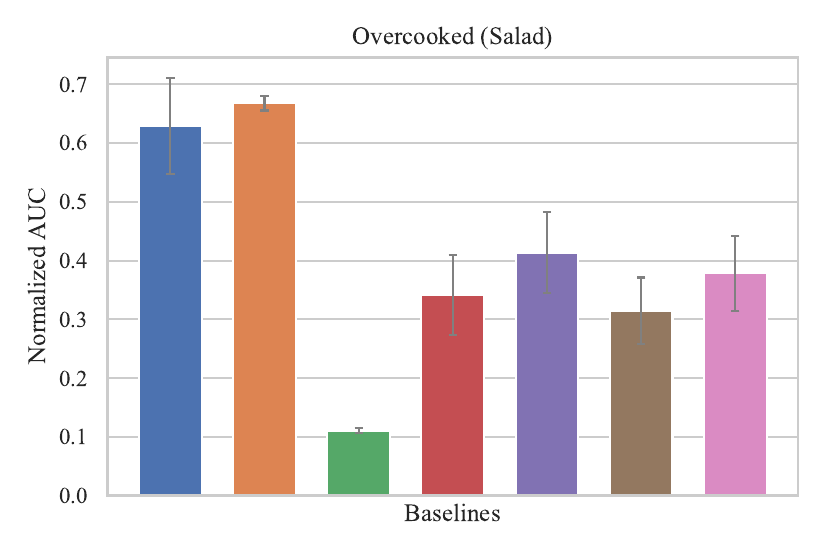}}
        \subfigure[Entropy]{
    \label{fig:overcookentropy}    		\includegraphics[width=0.32 \linewidth]{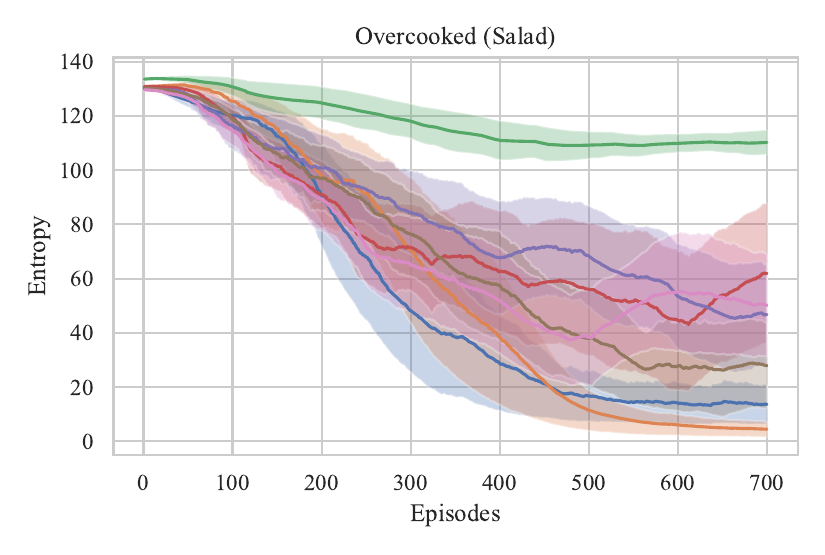}}

    \caption{ Results on Overcooked (Salad). (a) Training curves of baselines. We plot the average and standard error of rewards over $4$ seeds. (b) The average reward of baselines over all training episodes. (c) The entropy of action policy during training. 
    }
\label{fig:overcook}
\end{figure*}

In the Overcooked environment, as shown in \fig \ref{fig:overcook}, $\ourmethod$ exhibits the lowest action policy entropy demonstrates superior performance at the end of the training period. Similar to the previous environments, $\ourmethod$ shows its potential in mastering this game.

\end{document}